# De-identification of clinical free text using natural language processing: A systematic review of current approaches

Aleksandar Kovačević[1], Bojana Bašaragin[2], Nikola Milošević[2,3], Goran Nenadić[4]

[1] Faculty of Tehnical Sciences, Novi Sad, Republic of Serbia

[2] The Institute for Artificial Intelligence Research and Development of Serbia, Novi Sad, Republic of Serbia

[3] Bayer A.G., Berlin, Germany

[4] School of Computer Science, University of Manchester, Manchester, UK

**Abstract**

**Background:** Electronic health records (EHRs) are a valuable resource for data-driven medical research. However, the presence of protected health information (PHI) makes EHRs unsuitable to be shared for research purposes. De-identification, i.e. the process of removing PHI is a critical step in making EHR data accessible. Natural language processing has repeatedly demonstrated its feasibility in automating the de-identification process.

**Objectives:** Our study aims to provide systematic evidence on how the de-identification of clinical free text has evolved in the last thirteen years, and to report on the performances and limitations of the current state-of-the-art systems. In addition, we aim to identify challenges and potential research opportunities in this field.

**Methods:** A systematic search in PubMed, Web of Science and the DBLP was conducted for studies published between January 2010 and February 2023. Titles and abstracts were examined to identify the relevant studies. Selected studies were then analysed in-depth, and information was collected on de-identification methodologies, data sources, and measured performance.

**Results:** A total of 2125 publications were identified for the title and abstract screening. 69 studies were found to be relevant. Machine learning (37 studies) and hybrid (26 studies) approaches are predominant, while six studies relied only on rules. Majority of the approaches were trained and evaluated on public corpora. The 2014 i2b2/UTHealth corpus is the most frequently used (36 studies), followed by the 2006 i2b2 (18 studies) and 2016 CEGS N-GRID (10 studies) corpora.

**Conclusion:** Earlier de-identification approaches were mainly rule and machine learning hybrids with extensive feature engineering and post-processing, with more recent performance improvements are due to feature-inferring recurrent neural networks. Current leading performance is achieved using attention-based neural models. Recent studies report state-of-the-art F1-scores (over 98%) when evaluated in a particular manner adopted by the community. However, their performance needs to be more thoroughly assessed with different measures to judge their reliability to safely de-identify data in a real-world setting. Without additional manually labelled training data, state-of-the-art systems fail to generalise well across a wide range of clinical sub-domains.

# 1 Introduction

Electronic health records (EHRs) are a valuable resource for data-driven research in various areas [1], [2], [3]. However, the presence of personally identifiable information, often referred to as protected health information or PHI, makes EHRs unsuitable to be shared for research purposes. Information security mandates that the information which can be used to identify the patient should not be shared outside the clinical caregiving team without the explicit consent of the patient unless there are certain exceptional circumstances [4]. Therefore, removing PHI is a critical step in making EHR data accessible. Besides structured data, EHRs often contain free-text entries such as patient notes, admission and discharge summaries. Free-text de-identification is the process of detecting and replacing PHI in unstructured clinical text. The detected PHI can be replaced with a generic marker (e.g. 'NAME') or with realistic surrogates. The protected health information that needs to be replaced in the de-identification process is well-defined and often regulated by laws. In the US it is regulated by HIPPA (Health Insurance Portability and Accountability Act), in Canada by PHIPA (Personal Health Information Protection Act), and in the EU by GDPR (General Data Protection Regulation) [5]. These acts and regulations also define what is considered PHI. For example, HIPPA defines the Safe Harbour method, under which PHI includes eighteen types of identifiable information [6], some of them being names, dates (except year), telephone numbers, geographic data (e.g. address), fax numbers, social security numbers, email addresses, medical record numbers, account numbers, etc.

Manual de-identification is both time and labour-intensive and therefore, very costly [7], [8]. The median cost per single PHI can range between $0.71, for annotations by the initial annotator, to $377 for annotations by the forth annotator [7]. Therefore, there has been a need to automate and make this process cost effective by using natural language processing (NLP) . While some approaches have been trying to extract sentences that do not contain identifiable information by measuring frequencies of sentences and terms [6], [9], or by creating representations of clinical records that cannot be re-identified [10], [11], de-identification is usually modelled as a named entity recognition (NER) task. The most recent systematic review of systems that de-identify clinical free-text using NER has been performed by Meystre et al. in 2010 [12]. Since then, two shared tasks [13], [14] have been held, and significant methodological breakthroughs have been made through the use of deep learning [15]–[20]. For these reasons, our main aim is to provide a systematic review on how the methodologies in de-identification of clinical free text have evolved in the last thirteen years and to report on the performances and limitations of the current state-of-the-art systems. In addition, we aim to identify challenges and potential research opportunities in this field.

The remainder of the paper is organized as follows. The methodology of the review is described in detail in Section 2. Our findings are presented and discussed in Section 3. Finally, in Section 4, we conclude the review by outlining future research directions.

# 2 Methods

## 2.1 Overview

Following the guidelines described by Kitchenham [21], we devised our methodology around the following steps. First, we defined the research questions (RQs) to establish the scope, depth, and aim of the review. Next, we formed a search strategy to identify all relevant studies in a reproducible and efficient way. We further refined the scope of the review by defining the inclusion and exclusion criteria. A critical review of the included studies was then conducted to ensure the validity of the findings. We further extracted relevant information from each study to organise and synthesise evidence to support our findings.

## 2.2 Research Questions

The topic of this review is related to the properties of NLP approaches to de-identification of clinical free text. The research questions we aimed to answer with this review are given in Table 1. With RQ 1, we want to summarise the characteristics of methodologies applied in clinical de-identification, identify major methodological directions, and analyse how they evolved. RQ 2 aims at discovering the quality of de-identification systems, determining if they can be used to reliably de-identify data and whether they generalise well on different clinical domains. Finally, RQ 3 is focused on any potential challenges and opportunities in this field.

**Table 1. Research questions.**

| | **Research question** |
|---|---|
| **RQ 1** | Which methodological approaches have been used in de-identification of clinical free text? |
| **RQ 2** | What is the performance of clinical de-identification systems, and are they able to generalise well on corpora from different clinical domains? |
| **RQ 3** | What are the outstanding challenges in the field? |

## 2.3 Search Strategy

We used PubMed[1], Web of Science[2] and DBLP[3] computer science bibliography databases to retrieve relevant studies. PubMed is an interface to the National Library of Medicine's Medical Literature Analysis and Retrieval System Online (MEDLINE) database of biomedical literature and online books, with more than 35 million entries. Web of Science provides access to multiple databases with citation data form over 24,100 peer-reviewed journals, and various conference proceedings and books. It was initially produced by the Institute for Scientific Information (ISI) and is now maintained by Clarivate Analytics. DBLP was created by the University of Trier and Schloss Dagstuhl and provides bibliographic information on computer science journals and proceedings, including more than four million bibliographic units. PubMed and Web of Science enable searching titles and abstracts, while DBLP only implements search in titles, venue description and authors. PubMed entries can also be searched with terms from Medical Subject Headings (MeSH) controlled vocabulary.

[1] https://www.ncbi.nlm.nih.gov/pubmed/
[2] https://www.webofknowledge.com/
[3] https://dblp.uni-trier.de/

We designed our search query based on the Cochrane Handbook's search key points [22]. The query was based on three term clusters (see below). Full description of the search strategy and queries is given in Appendix A. Here, we provide brief descriptions of the three clusters used in the query:

1. **Topic**: Terms in this cluster are related to the topic of our review. MeSH contains only one term directly referring to our topic ("Data Anonymization"). However, a quick manual inspection revealed that several papers from the two de-identification shared tasks were not indexed with this term. Therefore we expanded the cluster with additional keywords, including "de-identification", "anonymization", "masking", "redaction" etc. We also used terms referring to two de-identification related shared tasks ("i2b2/UTHealth" and "CEGS N-GRID") held since 2010 (we restricted our review to a specific period – see Section 2.4 ).

2. **Domain**: With the second cluster, we restricted the query results to the clinical domain. The query comprised a number of keywords and abbreviations, such as "clinical", "medical", "EHR", "EMR", and similar. Since the term "clinical" is not included in MeSH we used the term "Electronic Health Records" as the most appropriate substitute.

3. **Method**: This cluster refers to NLP methods used to de-identify free text. The query included MeSH descriptors such as "Natural Language Processing", "Information Extraction", and "Machine Learning" along with several abbreviations and synonyms.

The term clusters were combined with the Boolean operator "AND", while the terms in each cluster were combined using the Boolean operator "OR". The searches were conducted in February 2023, and included the papers published by January 31, 2023.

## 2.4 Eligibility

As Meystre et al. published a systematic review with the same topic as ours in 2010 [12], we restricted our search from 2010 onwards[4]. Search results were screened in two stages. In the first stage, the following inclusion criteria were applied to titles and abstracts:

1. Must be a peer-reviewed journal article, conference or workshop paper, and the full text must be available.

2. Must be written in English.

3. Must be used for de-identification of clinical text written in the English language.

4. Must use real-world clinical data in the evaluation.

In the second stage, we retrieved full texts of the selected studies and tested them against the same criteria. We excluded studies if they did not describe a de-identification approach or a system or where the de-identification method was part of a larger framework but not evaluated separately. Two reviewers (AK, BB) conducted the screening in the first stage, while the second stage was based on an agreement between all four authors.

[4] Their paper was submitted in February 2010, and does not include relevant studies after that date.

## 2.5 Data Extraction

The following information was extracted from each selected study:

- Bibliographic information.
- Natural language processing approach (e.g. rule-based, machine learning, hybrid, etc.) along with the features used (e.g. lexical, syntactic, orthographic, etc.).
- Performance measurement (Precision, Recall, F-score) of the best performing approach (if different approaches have been evaluated, e.g., baselines).

All of the studies were thoroughly read. Some characteristics, such as features, were stored as categorical data, while others (e.g., a particular machine learning method used), were stored as directly quoted data. Details for each reviewed study are given in Appendix B.

# 3 Results and Discussion

The search in PubMed, Web of Science and DBLP retrieved 828, 369, and 1,171 studies, respectively. After duplicates were removed (243 studies), 2,125 titles and abstracts were screened in the first stage. As a result, 112 studies were selected for full-text screening, and 69 studies were included in the review. The review flow diagram is presented in Figure 1.

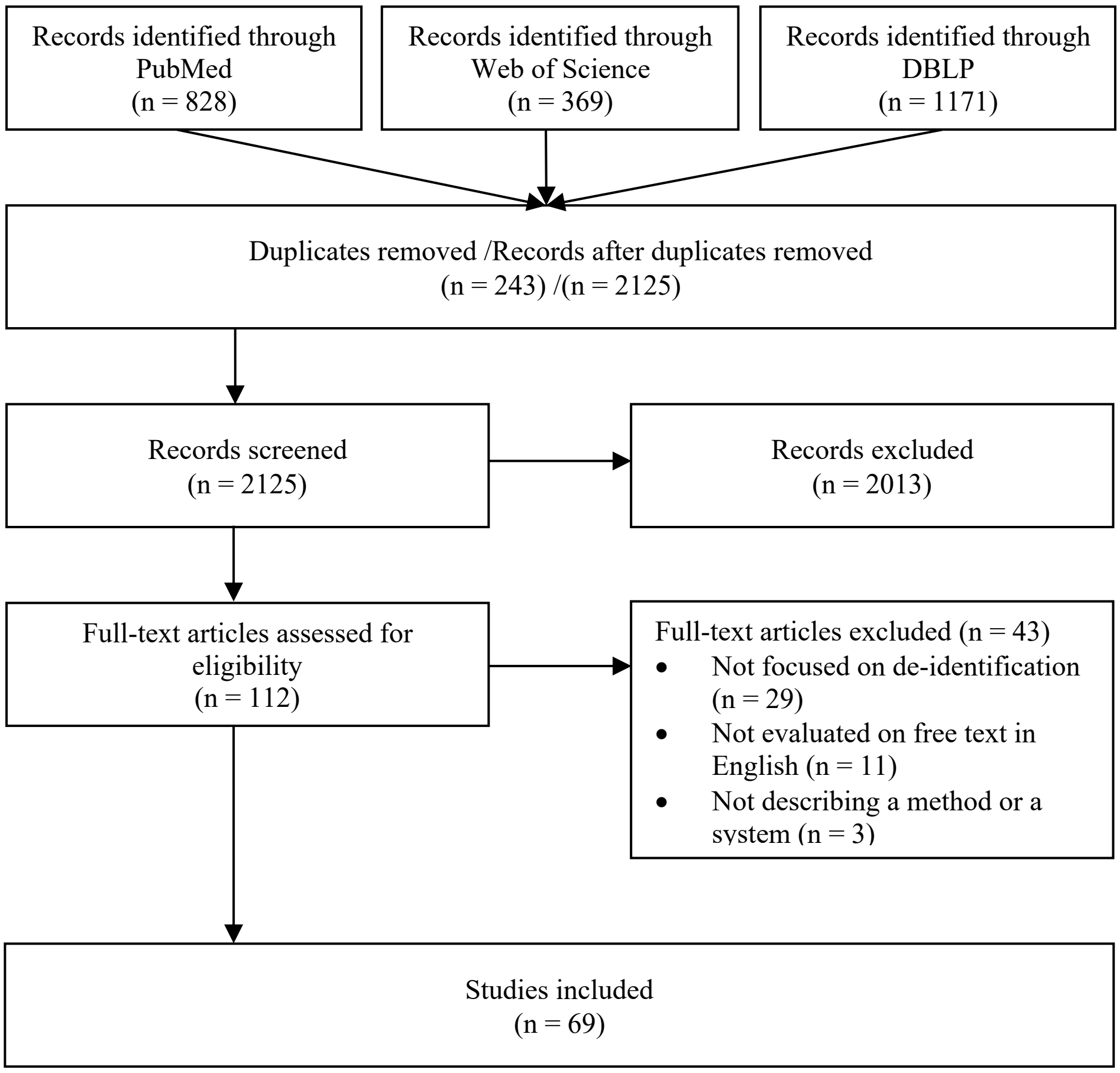

**Figure 1 Review flow diagram.**

## 3.1 Publication Demographics

Majority of the reviewed studies are published in the Journal of Biomedical Informatics (12 studies), BMC Medical Informatics and Decision Making (7 studies), Journal of the American Medical Informatics Association (7 studies), the Proceedings of the American Medical Informatics Association Annual Symposium (6 studies), and the International Journal of Medical Informatics (4 studies). Researchers from the United States (39 studies), China (8 studies), and the United Kingdom and Canada (5 studies each) have authored over 80% of the reviewed publications, and the rest was contributed by the authors from Australia, India, Ireland, Russia, Italy, Spain, Japan, Serbia, and Vietnam. The majority of the studies were published in 2017 (11 studies)2022 (8 studies), and 2015, 2016, and 2020 (7 studies) following the 2014 i2b2/UTHealth and 2016 CEGS N-GRID shared tasks.

## 3.2 Corpora

De-identification approaches before 2010 have been developed and evaluated mostly on proprietary corpora [12]. The only exceptions were systems dedicated to the 2006 i2b2 shared task [23]. Between 2010 and 2015, eight out of twelve studies still relied on proprietary corpora (Figure 2). After the i2b2/UTHealth shared task was held in 2014 [14], and its resources released in 2015, publicly available corpora became dominant (Figure 2). The 2014 i2b2/UTHealth is the most frequently used corpus in clinical de-identification research (36 studies). The researchers also favoured the corpora from the other two shared tasks, 2006 i2b2 (18 studies) and 2016 CEGS N-GRID (10 studies).

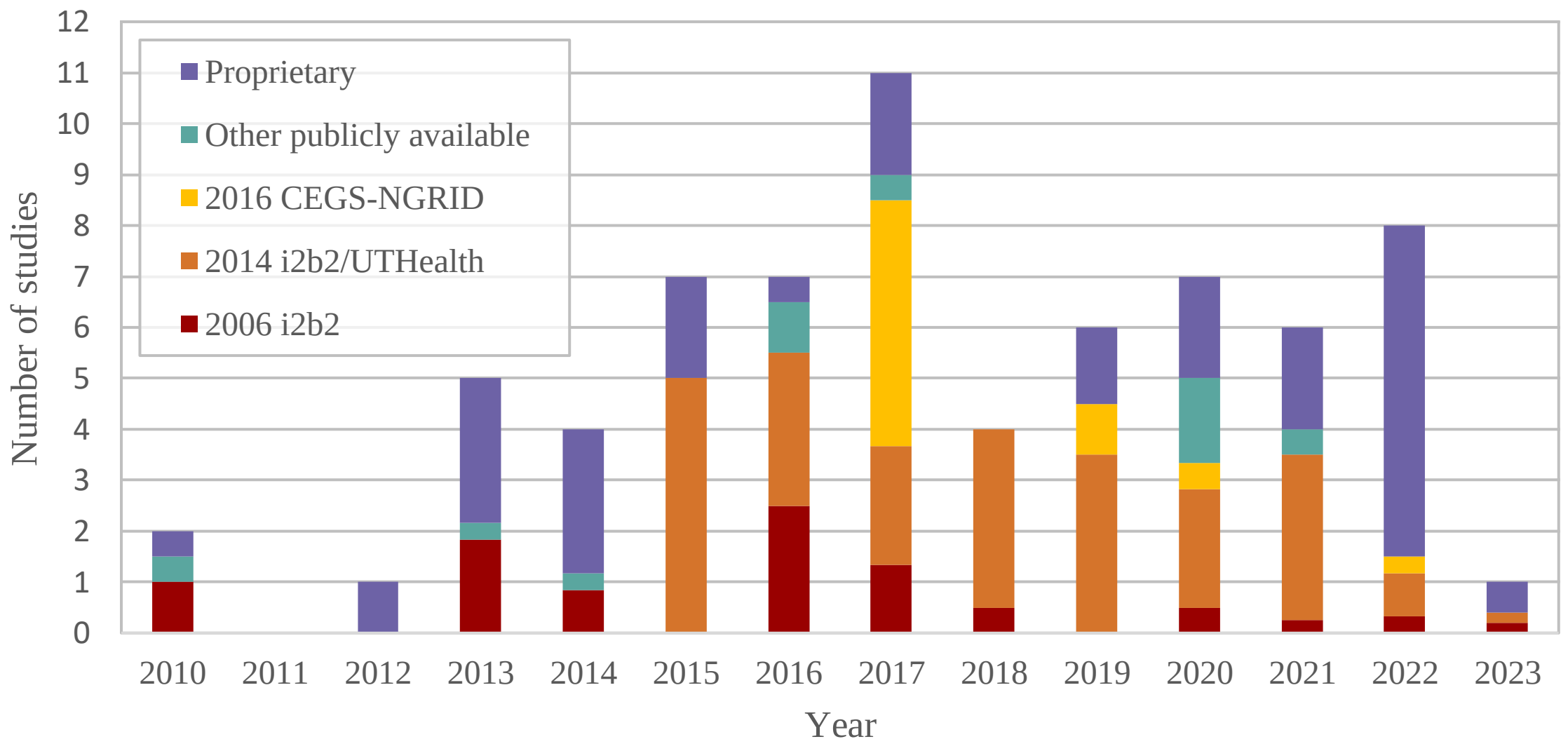

**Figure 2 The distribution of studies per year and corpora.**

After 2020 researchers started turning to proprietary corpora again since the inability to share more varied clinical text started impeding progress in the medical domain [19]. Table 2 provides a summary of each corpus used in the studies. Below we provide detailed descriptions of the three most frequently used corpora.

**Table 2. Description of corpora used for de-identification of clinical free text.**

| Dataset | Provenance | Documents | Size (train:test) | Annotator agreement (metric) | Studies |
|---|---|---|---|---|---|
| **2006 i2b2** [23] | Partners Healthcare (PHC) | Discharge summaries | 889 (669:220) | - | [18], [19], [24]–[38] |
| **2014 i2b2/UTHealth** [39] | Partners Healthcare (PHC) | Longitudinal medical records | 1,304 (790: 514) | 89.50% (F1-Score) | [10], [15]–[20], [27], [28], [34], [37], [38], [40]–[59] |
| **2016 CEGS N-GRID** [13] | Partners Healthcare (PHC) | Psychiatric intake records | 1,000 (600: 400) | 85.10% (F1-Score) | [17], [38], [45], [46], [53], [60]–[63] |
| **PhysioNet (MIMIC-II)** [64]–[66] | Beth Israel Deaconess Medical Center (BIDMC) | Nursing notes | 2,434 (1,436:998) | - | [19], [25]–[27], [40] |
| **MTSamples** [35] | www.mtsamples.com | Clinical notes | 1,885 | - | [35] |
| **MIMIC-III-D** [15] | Beth Israel Deaconess Medical Center (BIDMC) | Discharge summaries | 1,635 (1308:327) | - | [15], [19], [67] |
| **MIMIC-III-T** [68] | Beth Israel Deaconess Medical Center (BIDMC) | Discharge summaries | 4,441 (80%:20%) | - | [40] |
| **MIMIC-III-H** [27] | Beth Israel Deaconess Medical Center (BIDMC) | Discharge summaries | 1,635 (1,308:327) | - | [27] |
| **VHA** [69] | Veterans Health Administration (VHA) | Consult notes | 800 | 83.00% (-) | [36], [69] |
| **CareWeb** [70] | University of Michigan Health System (UMHS) | History and physical notes and social work notes | 360 (288: 72) | - | [70] |
| **SLaM** [71] | South London and Maudsley NHS Trust (SLaM) | Patient notes | 500 | - | [71] |
| **CCHMC** [25] | Cincinnati Children's Hospital Medical Center (CCHMC) | Clinical notes | 250 | 91.76% (F1-score) | [25] |
| **VUMC** [29] | Vanderbilt University Medical Center (VUMC) | Discharge summaries, laboratory reports, letters, order summaries | 500 | - | [29] |
| **CINSW** [35] | Cancer Institute New South Wales (CINSW) | Pathology and cytology reports | 852 | - | [35] |
| **NIH** [72] | NIH Clinical Center (NIHCC) | Clinical notes and radiology reports | 3,093 (1140:1953) | - | [72] |
| **VA** [73] | Department of Veterans Affairs (VA) | Clinical notes | 473 | - | [73] |
| **NORC** [74] | Nutrition Obesity Research Center (NORC) at the University of Chicago | Medical transcriptions | 1429 | - | [74] |
| **CNHS** [75] | Christie NHS Foundation Trust (CNHS) | Clinical notes | 1300<br>(780:520) | - | [75] |
| **CPFT** [4] | Cambridgeshire and Peterborough NHS Foundation Trust (CPFT) | Clinical notes | 100 | - | [4] |
| **TAMHSC** [76] | Texas A&M Health Science Center | Clinical notes | 500 | - | [76] |
| **UTHealth** [63] | University of Texas Health Science Center at Houston (UTHealth) | Outpatient notes | 325 | - | [63] |
| **UFHealth** [52] | University of Florida Health (UFHealth) | Clinical notes | 500<br>(300:200) | 88.90%<br>(-) | [52] |
| **USHP** [77] | US-based healthcare provider | Clinical notes | 400 | - | [77] |
| **AHO** [78] | Anonymous healthcare organization | Clinical study reports | 370 | - | [78] |
| **ICES** [11] | Institute for Clinical Evaluative Sciences | Consult notes | 9051707 | - | [11] |
| **UPHS-S** [79] | University of Pennsylvania Health System (UPHS) | Radiology reports | 2,501/2,503<br>(1,480:1,023) | 0.938<br>(Cohen's k) | [18], [79] |
| **MIMIC-III-J** [80] | Beth Israel Deaconess Medical Center (BIDMC) | Radiology notes in ICUs, Discharge notes in ICUs [37]<br>Progress notes [11] | 1,000 sampled each [37],<br>2,6 million [11] | - | [11], [37] |
| **RIH** [57] | Rhode Island Hospital (RIH) | Health record notes | 25 | Unannotated (used for qualitative assessment) | [57] |
| **HCR** [81] | - | Hybrid clinical reports: Consult notes, on-treatment visits, phone encounters, treatment records, follow-ups | 5,200 | - | [81] |
| **MC** [58] | Mayo Clinic | Physician notes, medicine administration records | 104 million<br>10,000 sampled | 0.9694<br>(Cohen's k) | [58] |
| **UMH-R** [82] | University of Missouri Healthcare (UMH) | 7 types of radiology reports | 10,239<br>(75%:25%) | - | [82] |
| **AHDS** [20] | two principal referral hospitals in Sydney, Australia | Hospital discharge summaries | 3,554<br>600 annotated<br>(400:100:100) | 96.65<br>(Cohen's kappa, all tokens)<br>92.55<br>(Cohen's kappa, annotated tokens)<br>94.74<br>(F1 strict entity) | [20] |
| **KUMC/MCW** [83] | The University of Kansas Medical Center (KUMC), The Medical College of Wisconsin (MCW) | Hospital discarge summaries (22 types of patient records) | 48 million<br>(:48,000) | - | [83] |
| **IC** [84] | Internal corpus | Discharge summaries | 500<br>(70%:10%:20%) | - | [84] |
| **UMH-P** [85] | University of Missouri Healthcare (UMH) | (NSCLC) Pathology reports | 1500<br>(3:1 ratio) | - | [85] |
| **POWCC/POWCC MOSAIQ** [86] | Prince of Wales Hospital Cancer Centre (POWCC), POWCC MOSAIQ (Elekta, Stockholm, Sweden) OIS | Oncology reports | 52,716<br>300 sampled | - | [86] |
| **UPHS-C** [59] | University of Pennsylvania Health System (UPHS) | Clinical encounter notes | 14,828,230 | - | [59] |
| **UPHS-R** [18] | University of Pennsylvania Health System (UPHS) | Radiology reports | 999 | - | [18] |
| **SMC** [87] | Stanford medical center (SMC) | Radiology reports | 500<br>(:500) | - | [18] |

The corpus used for the Informatics for Integrating Biology and the Bedside (i2b2) de-identification challenge held in 2006 comprises 889 discharge summaries from Partners Healthcare (PHC) [23]. The summaries were annotated with eight PHI categories: six required by HIPPA (*Patient*, *Location*, *Date*, *ID*, *Phone number* and *Age*) and two added by the organizers (*Doctor* and *Hospital*). Each summary was manually labelled by three annotators (undergraduate and graduate students and a professor).

The 2014 i2b2/UTHealth corpus includes longitudinal records for 296 patients, with 2-5 records per patient, obtained from PHC [39]. The annotation scheme is based on HIPPA and additionally expanded with information that can indirectly identify patients, such as all parts of dates, including years, locations (including states and countries) information about hospitals, doctors and nurses, and patient's professions. The full list of annotated categories is given in Table 3. Both token and entity level inter-annotator agreement (IAA) was high (micro F-measures of 0.930 and 0.895, respectively).

The 2016 CEGS N-GRID corpus contains raw text of 1000 psychiatric intake records obtained from Partners Healthcare [13]. The records were annotated with the categories presented in Table 3. The IAA was slightly lower than for the 2014 i2b2/UTHealth corpus (micro F-measures of 0.851 and 0.913 for token and entity level, respectively). In contrast to the 2014 i2b2/UTHealth corpus, psychiatric intake records contain a significant amount of spelling and punctuation errors and considerably more information that could be used to identify the patients, e.g. places lived, jobs held, hobbies, traumatic events, even pet names etc. [13]. In each of the three corpora (2006 i2b2, 2014 i2b2/UTHealth and 2016 CEGS N-GRID) PHI was replaced with realistic surrogates.

**Table 3 Classification of PHI into categories and sub-categories, as defined in the 2014 i2b2/UTHealth corpus – adapted from [15].**

| PHI categories | PHI sub-categories | Description | Required by HIPPA |
|---|---|---|---|
| **AGE** | AGE | Ages ≥ 90 | ✓ |
| | | Ages ≤ 90 | |
| **CONTACT** | PHONE | Telephone numbers | ✓ |
| | FAX | Fax numbers | ✓ |
| | EMAIL | Electronic mail addresses | ✓ |
| | URL | Uniform resource locators | ✓ |
| | IP ADDRESS | Internet protocol addresses | ✓ |
| **DATE** | DATE | Dates (month and day parts) | ✓ |
| | | Year | |
| | | Holidays | |
| | | Days of the week | |
| **ID** | IDNUM | Social Security numbers | ✓ |
| | | Account numbers | ✓ |
| | | Certificate or license numbers | ✓ |
| | MEDICAL RECORD | Medical record numbers | ✓ |
| | DEVICE | Vehicle or device identifiers | ✓ |
| | HEALTH PLAN | Health plan numbers | ✓ |
| | BIOID | Biometric identifiers or full-face photographs | ✓ |
| **LOCATION** | STREET | Street address | ✓ |
| | CITY | City | ✓ |
| | ZIP | Zip code | ✓ |
| | STATE | State | |
| | COUNTRY | Country | |
| | LOCATION-OTHER | Other identifiable locations such as landmarks | |
| | ORGANIZATION | Employers | ✓ |
| | HOSPITAL | Hospital | |
| **NAME** | PATIENT | Names of patients and family members | ✓ |
| | DOCTOR | Provider name | |
| | USERNAME | User IDs of providers | |

| | |
|---|---|
| **PROFESSION** | Profession |

## 3.3 De-identification Methods

De-identification methods typically include the following stages: pre-processing, protected health information recognition and classification, and finally, optional post-processing (Figure 3). Some systems such as [36] perform named entity recognition and classification in two separate steps.

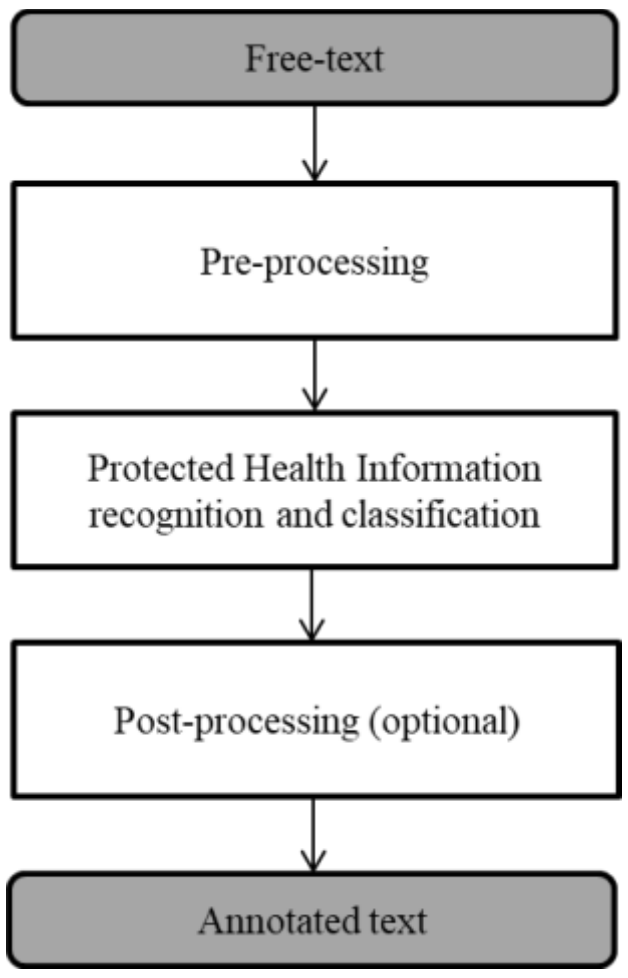

**Figure 3 A typical procedure for de-identification of clinical free text.**

Pre-processing prepares raw text for the next steps in the de-identification pipeline and includes tasks such as sentence splitting, tokenisation, part-of-speech tagging, and chunking. Surprisingly, general domain tools such as Stanford CoreNLP (10 studies), OpenNLP (5 studies), and GATE (4 studies) are more frequently used (see Table 4) than specialised biomedical tools, e.g., cTAKES (4 studies), CLAMP (2 studies), GENIA Tagger (2 studies), and MedEx (2 studies). He et al. [43] and Jiang et al. [62] found that specialised tokenizers improve the overall F1-score by 2-3%. Since clinical text often contains abbreviations, misspellings and conjoint words [13], [39], authors of 14 studies developed their specialised tokenisation modules.

**Table 4 Natural language processing tools used by the reviewed studies.**

| Name | URL | Studies |
|---|---|---|
| **Stanford CoreNLP** | https://stanfordnlp.github.io/CoreNLP/ | [16], [32], [33], [35], [44], [44], [50], [63], [76], [77] |
| **OpenNLP** | https://opennlp.apache.org/ | [43], [62], [63], [75], [77], [88] |
| **cTAKES** | http://ctakes.apache.org/ | [36], [42], [61], [69] |
| **GATE** | https://gate.ac.uk/ | [4], [42], [61], [75] |
| **CLAMP** | https://sbmi.uth.edu/ccb/resources/clamp.htm | [63], [88] |
| **dTagger** | https://lsg3.nlm.nih.gov/LexSysGroup/Projects/dTagger/current/index.html | [72], [89] |
| **GENIA Tagger** | http://www.nactem.ac.uk/GENIA/tagger/ | [43], [51] |
| **Lucene** | https://lucene.apache.org/ | [36], [69] |
| **MedEx** | https://code.google.com/p/medex-uima/downloads/list | [4], [16] |
| **SpaCy** | https://spacy.io/ | [10], [37], [45], [55], [59], [79] |
| **MetaMap** | https://metamap.nlm.nih.gov/ | [29], [53] |
| **NLTK** | http://www.nltk.org/ | [48], [59], [81] |
| **SENNA** | https://ronan.collobert.com/senna/ | [63] |
| **TreeTagger** | https://www.cis.uni-muenchen.de/~schmid/tools/TreeTagger/ | [25] |
| **WordNet** | https://wordnet.princeton.edu/ | [67] |

Methods for de-identification can be classified into three broad categories: rule-based, machine learning-based or hybrid. Shared tasks provided a boost for methodological developments of de-identification. Over 60% of the systems were either developed during the tasks or used the released corpora. Figure 4 illustrates the distribution of methodologies from 2010 to 2023.

**Rule-based systems:** About a third of the studies (31), applied at least one rule-based tagger, however, only six reviewed studies were entirely rule-based in contrast to the period before 2010, when over half of the studies were rule-based [12]. PHI categories that are mostly detected by rule-based methods are contact information (22 studies), person names (19 studies), dates (18 studies), and geographical locations (18 studies). The rule-based systems mainly relied on regular expressions [4], [58], [72], [73], [76], [83], [86], a set of heuristics [71], [72] or utilization of structured fields from electronic health records (EHRs) [71], [4], [58], [76], [83].

**Machine learning-based systems** can be divided into traditional and those based on deep learning. The traditional systems rely on a set of hand-crafted features extracted from the training corpus. While systems with hand-crafted features were used before 2010, a significant increase in the use of these systems can be noted in the years ahead. Among the traditional ML-based systems, Conditional Random Fields (CRF) [90] is the most widely applied model, used in twenty ML-based and eighteen hybrid systems. The hand-crafted features used in the ML studies can be divided into seven categories (Table 5).

**Table 5 Types of hand-crafted features used in machine learning-based methods.**

| Features | Description | Frequency |
|---|---|---|
| **Lexical** | Features that include character or token n-grams, n-grams of tokens in a context window and n-grams of lemma, stem, part-of-speech, etc. | **29/51** |
| **Dictionary** | A feature indicating whether a term matches one or more dictionary entries. We also include features extracted by external sources (i.e., NLP tools). | **23/51** |
| **Orthographic** | Orthographic patterns associated with a token, e.g., starting with a lowercase letter, containing only alphanumeric characters, containing symbols, etc. | **20/51** |
| **Positional** | Position of a token relative to the sentence or document. | **11/51** |
| **Rules** | The use of rules as features. | **9/51** |
| **Word** | Word representation features based on word2vec, Brown clustering etc. | **27/51** |
| **representation** | | |
| **Semantic** | Features indicating meaning in the language and the relationships among the words. | **9/51** |

The choice of hand-crafted features and ML models per study can be seen in Table 6. Lexical (29 studies), word representation (27 studies), and dictionary (23 studies) are the most frequently extracted feature groups. Studies after 2020 make an extensive use of word representations. Twelve studies evaluated the impact of feature groups on the overall performance. Lexical features were most beneficial in five studies [16], [26], [35], [43], [88]. Dictionary, rule-based, and orthographic features seemed to be the most influential in two studies [31], [62]. Authors of three studies, [38], [82], [85], reported that a combination of lexical, orthographic, and word representation features was most beneficial in their study. Lee et al. [67] and Abu-El Rub et al. [83] found that features obtained from structured EHR entries increase the recall, especially for person names. Murugadoss et al. [58] used patient-specific information (e.g. names, addresses) from structured EHRs to augment their model training and match against free-text PH.

**Table 6 Hand-crafted features used in ML-based and hybrid approaches.**

| Study | Features | ML model | Study | Features | ML model |
|---|---|---|---|---|---|
| [26] | Lexical, Dictionary, Orthographic, Rules. | CRF | [67] | Dictionary, Rules, Word representation | LSTM |
| [36] | Lexical | CRF, SVM | [75] | Lexical, Dictionary, Orthographic | CRF |
| [69] | Lexical, Dictionary, Orthographic | CRF, SVM | [33] | Lexical, Positional | CRF |
| [31] | Lexical, Dictionary, Rules, Semantic | J48 | [46] | Lexical, Dictionary, Orthographic, Semantic, Word representation | CRF, LSTM |
| [25] | Lexical, Dictionary, Orthographic | CRF | [61] | Lexical, Dictionary, Orthographic | CRF |
| [35] | Lexical, Orthographic, Rules, Positional | CRF | [88] | Lexical, Dictionary, Orthographic, Positional, Semantic | CRF |
| [43] | Lexical, Dictionary, Orthographic | CRF | [62] | Lexical, Dictionary, Orthographic, Semantic | CRF, LSTM |
| [16] | Lexical, Dictionary, Orthographic, Positional, Word representation | CRF | [63] | Lexical, Dictionary, Positional, Semantic | CRF |
| [51] | Lexical, Dictionary  Rules | CRF | [30] | Lexical, Dictionary, Positional | CRF |
| [42] | Lexical, Dictionary, Orthographic, Positional | CRF | [34] | Dictionary, Semantic, Word representation | GRU |
| [41] | Lexical, Dictionary, Orthographic | HMM, CRF | [91] | Lexical, Dictionary, Orthographic, Positional | CRF |
| [74] | Rules, Semantic | Logistic reg. | [78] | Lexical | CRF |
| [32] | Lexical, Positional | CRF | [92] | Lexical, Dictionary, Rules | LSTM |
| [48] | Lexical, Dictionary, Word representation | CRF, LSTM | [27] | Dictionary, Ortographic | LSTM |
| [10] | Word representation | DANN | [15], [20], | Word representation | LSTM |
| | | | [47], [55], [57] | | |
| [28], [45] | Word representation | LSTM, GRU | [49], [50] | Word representation | RNN |
| [81] | Lexical, Word representation | MLP | [93] | Lexical, Dictionary, Rules | LSTM |
| [44], [53] | Word representation | Ensemble of 12 de-ide methods (CRF, LSTM, SVM) | [11], [59] | Word representation | CBOW, Skipgram (Logistic regression, SVM, CNN) |
| [40] | Word representation | GRU, stacked RNNs, self-attention | [79] | Orthographic, Word representation | Emory HIDE, NeuroNER |
| [37] | Orthographic, Word representation | LSTM | [85] | Lexical, Orthographic, Positional, Rules, Word representation | CRF |
| [82] | Lexical, Orthographic, Positional, Rules, Semantic, Word representation | CRF | [38] | Lexical, Orthographic, Word representation | LSTM |
| [84] | Semantic, Word representation | LSTM | | | |

Following the release of the 2014 i2b2/UTHealth corpus and the growing popularity of deep learning, many approaches applied recurrent neural networks (RNN) to the de-identification task. Dernoncourt et al. were the first to apply the LSTM network in clinical free-text de-identification [15]. They achieved results comparable to the best-performing feature engineering hybrid system (Yang and Garibaldi [51]). Ever since, researchers experimented with the GRU networks [34], [45], [40], DANN networks [93], different embeddings [17], [45], [48], [11], [59] addition of hand-crafted [67] or text skeleton features [34], perceptron [81], and transformer architectures [18], [19], [40], [54], [56]–[58]. The most popular RNN architecture in the reviewed studies is the bidirectional LSTM network with a CRF inference layer (Appendix D). Interestingly, most of the studies relied on embeddings trained on general domain corpora. Some studies even reported that using word embeddings trained on clinical or biomedical corpora provides no advantage [15], [19], [46], [59]. Contrary to those findings, Syed et al. [84] reported that using Flair mixed-domain clinically pre-trained embeddings made a statistically significant improvement to F1-score of 1.05%. Along those lines, Abdalla et al. [11], who used Pearson correlation to compare the embeddings created using the obfuscation anonymization technique and embeddings on out-of-domain corpora, claimed that the former remain more informative. GloVe (10 studies) and word2vec (8 studies) and are most frequently used embedding methods.

Most of the effort put into feature engineering earlier was then spent on tuning various RNN hyper-parameters that have significant impact on the performance. For example, Jiang et al. found that optimising the value of the text window size (used as input for the RNN) can improve the overall F1-score by 3% [62]. The performance of the RNN models could be further boosted by deep contextualised embeddings [17], [20], [37], [38], [45], [84], using hand-crafted features [46], [67], or integrating them with the CRF, rules, and other models [15], [44], [46].

Transformer-based architectures, namely BERT and its fine-tuned versions – SciBERT and BioBERT, were first used for the task of clinical free-text de-identification in 2020 [19]. Since then, DistilBERT [54], PubMedBERT [18], BERT-like [40] or general transformer architectures [57] were used to achieve state-of-the-art results.

**The hybrid systems** were dominant until 2016. The typical hybrid de-identification system consists of a machine learning and a rule-based component. A machine learning component includes multiple single-class (or one multi-class) CRF models trained with an extensive set of hand-crafted features (Table 6). The rule-based component is usually realised through a set of regular expressions (developed using lexical features) that cover more formulaic categories such as dates, contact information, identifiers etc. Both components often make use of dictionaries that are manually or semi-automatically collected from external resources or training data. Hybrid models have produced state-of-the-art results (Appendix D), but achieving these results requires much engineering. For example, Yang and Garibaldi generated 220 million features from their ML model and used a post-processing component with eleven sub-steps [51]. Post-processing had a significant impact on the final results, increasing the F1-score by more than 20% for some categories. Similarly, with additional pre-processing, new features, and feature selection He et al. improved their official shared task F1-score by 4% [43].

A number of approaches used post-processing to improve the performance or to resolve of conflicts in resulting annotations in hybrid approaches [16], [25], [30]–[33], [36], [42], [46], [51], [60], [61], [69], [77], [83], [88]. The post-processing was performed by training an additional classifier to filter out false positives [36], [69] or by a set of heuristics [31], [32], [77], [42], [25], [51], [58], [61], [83]. Some of these approaches boosted F1-score up to 26% for certain PHI categories. [58] applied post-processing to improve the fidelity of PHI surrogates, matching the ethnicity, gender and date format to the original document.

As the Figure 4 shows, two primary methodological directions lately are machine learning, particularly deep learning-based, and a constant number of hybrid systems.

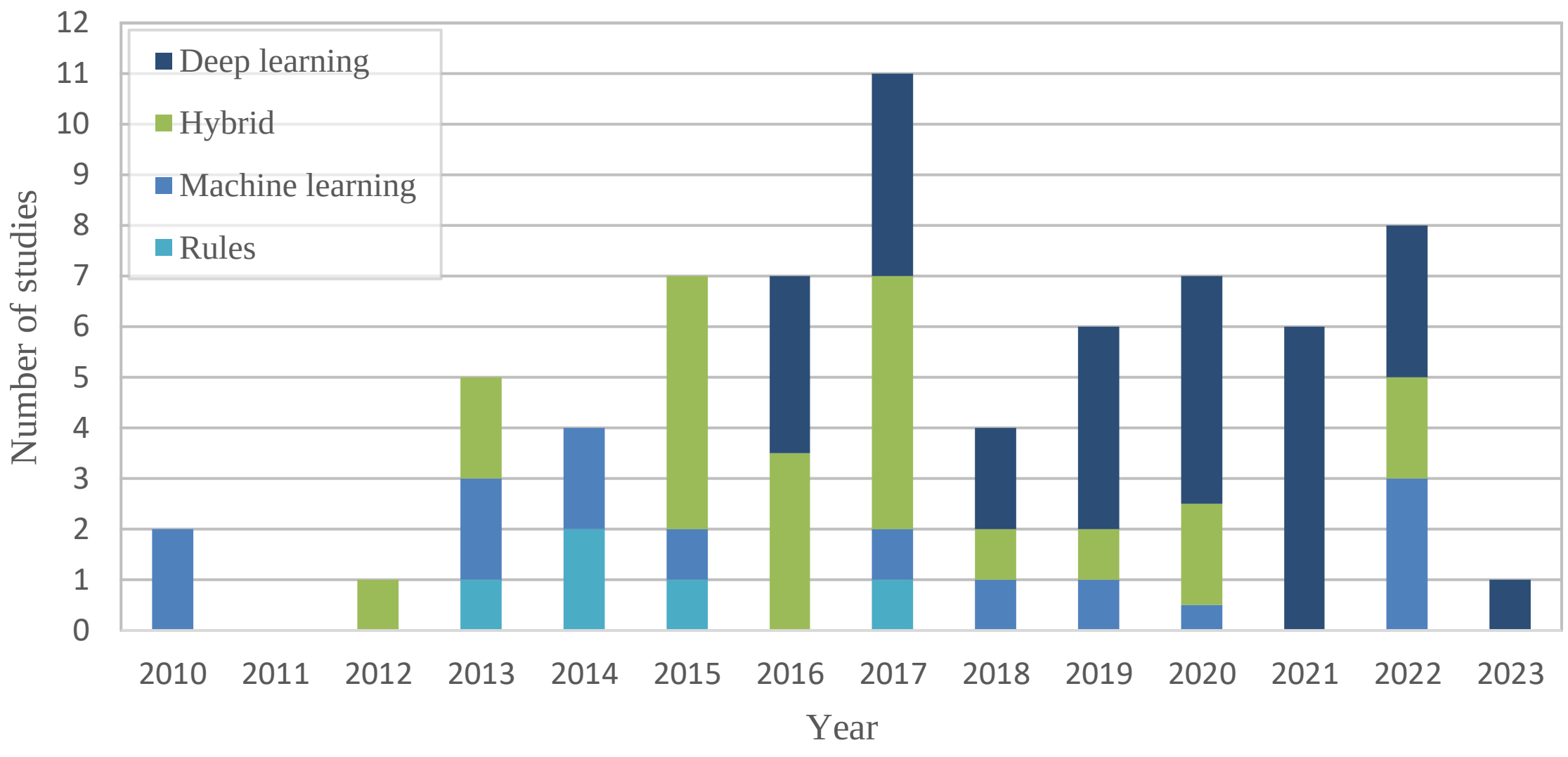

**Figure 4 The distribution of studies per year and method.**

## 3.4 Performances

The quality of a de-identification system can be evaluated from different aspects. It is of the utmost importance that PHI entities are detected successfully. To the best of our knowledge, there is no industry-wide standard. However, an F1-score of 95% has been suggested as a rule-of-thumb for determining whether a system can reliably de-identify a corpus for safe distribution [14]. It has also been suggested that binary token-level matching considering only HIPAA required categories is the most appropriate way to calculate the F1-score from the aspect of safety [15]. Almost all of the systems, especially the best-performing ones, evaluated on the most widely used publicly available corpora (Appendix D) provided token-level F1-scores (binary or otherwise) greater or equal than 95%. Previous reviews [12] and shared task overviews [13], [14] showed that de-identification at the 100% level of accuracy was still not feasible. Although our review confirms that perfect F1-scores are still out of reach, we do record studies that report achieving 100% recall using certain methods and on certain datasets. Abdalla et al. [11] developed a method that achieved 100% recall using word embeddings to replace every token in clinical data, albeit at the expense of readibility. Ahmed et al. [40] reached 100% recall on the MIMIC-III dataset using their proposed GRU network. Zhou et al. [81] reached 100% recall on two randomly selected reports using word and contextualized embeddings with a multilayer perceptron classifier. Still, most recent automated systems can identify PHI entities at the level that is considered reliable by the standard suggested in [14], e.g., some of the best-performing systems [17], [44] miss less than 2% of the PHI tokens (recall over 98%) at the precision level over 99%. However, performances at PHI category level, especially the differences between the micro and macro F1-scores reveal that further efforts should be put into standardising the way that PHI detection is evaluated. For example, the binary token HIPAA F1-score reported by Tang et al. [17] does not express that some of the HIPPA required categories (*Contact*, *Id*, *Location* and *Name*) are not consistently recognised with high F1-scores (Appendix E). The *Location* and *Name* categories especially in the case of the 2016 CEGS-NGRID corpora show that even with a large number of training instances and contextualised embeddings, state-of-the-art RNN-based systems fail to obtain F1-scores over 95% (Appendix E). The *Id* category in the 2016 CEGS-NGRID corpora is another example that shows that overall scores calculated at the micro-level can be misleading. Identifiers were scarce in the 2016 CEGS-NGRID corpora (only 38 examples in training and 21 in the test set), so ML models did not have enough training data (Appendix E). As the micro-level F1-score was the shared task performance ranking measure, even the participants with hybrid systems did not put much effort into developing rules for infrequent categories [88]. However, in a real-world setting all PHI categories should be addressed.

The primary goal of the de-identification of clinical documents is to facilitate medical research through the safe sharing of clinical documents. Thus, another critical aspect of the evaluation of de-identification systems is the usability of the de-identified documents. Meystre et al. used the term 'over scrubbing' to refer to clinical documents that were devoid of potentially useful information after de-identification [12]. They have also reported that none of the studies included in their review considered over scrubbing errors. However, several studies included in our review have focused on the impact of de-identification on further data use [11], [25], [40], [53]. Deleger et al. evaluated the performance of medication detection system on the original, manually, and automatically de-identified clinical notes from the Cincinnati Children's Hospital Medical Center [25]. They found that de-identification had no statistically significant impact on medication detection performance. The F1-score was even 0.2% higher on de-identified notes. Bui et al. used the 2014 i2b2/UTHealth corpus to train a CRF model [91] and experiment with single vs multi-class de-identification. Specifically, their goal was to determine if clinical documents have the same level of readably if all PHI are replaced with a single surrogate (e.g., "PHI"). They found that three expert reviewers reached a high agreement (Kappa of 0.975) that single and multi-class de-identification both have the same impact on readably. Abdalla et al. [11] evaluated the embeddings created by token obfuscation on tasks of ICES diagnostic code and MIMIC-III ICD-9 code classification, along with the task of sentiment analysis. While observing that the general trend is decreased performance, the decrease is slight (2% and 3%, in code classification and 8% in sentiment analysis). Ahmed et al. [40] introduced two utility metrics (BLEU and topic modeling) and performed ICD9 code classification on MIMIC-III. They reported an improvement in comparison with their baseline ([15]), as well as improved accuracy in the classification task, due to class-name simplifications in de-identified datasets. They concluded that contrary to relying on high recall in the de-identification task, both precision and recall need to be considered when analysing utility.

## 3.5 Generalisability

The ability to perform well on a diverse set of previously unseen corpora is the ultimate goal of every NER system. One of the tracks at the 2016 CEGS-NGRID shared task was aimed at evaluating the existing de-identification tools on the previously unseen corpus (2016 CEGS-NGRID corpus) without providing any training data [13]. The results demonstrated that the existing systems do not generalise well. Only two out of eight submitted systems provided an F1-score over 70% at the entity-level. The F1-score of the winning team was 79.85%. The same team achieved an F1-score of 91.43% when training data was available [46]. Our review confirmed that without customisation (using annotated examples from the target corpus), de-identification systems fail to generalise. Significant drops in F1-score were observed in multiple studies [19], [27], [35], [37], [52], [53], [63] when applying ML methods trained on one corpus to a different corpus. However, in their domain adaptation experiments of using pseudo-labels or just a few labeled examples from the target domain, Shun et al. [37] managed to achieve improvements in F1-score of betweeten 5.38 and 10.86. They found that the improvement is higher with a smaller amount of labeled data from the target set, e.g. as much as 7.99 for only 0.1% of labeled data. Using the pseudo-label approach, they managed to achieve strict name entity F1-scores of 56.71 to 86.35 in cross-domain experiments. Given that clinical free-text data contains local jargon and identifiers, and that machine learning models tend to overfit to at least some extent, these performance drops were not unexpected. For example, [27] observed that over 50% of false negatives (made by the model trained on the 2014 i2b2/UTHealth corpus) on the PhysioNet corpus were mainly local abbreviations for locations such as 'MICU', 'PMICU', 'cath lab', and similar. [19] attribute false negatives to the lack of standardization guidelines when annotating PHI. They also noted that a high capacity model (such as BERT large) can deliver over 95% recall in cross-domain experiments, even for scarcely labeled entities. The embedding approach of Abdalla et al. [11] reports promising results in terms of generalizability, showing that embeddings created from obfuscated clinical data can be used for other types of clinical corpora and even out-of-domain corpora (movie reviews and GoogleNews).

## 3.6 Opportunities and Challenges

**Increasing performance for lower-performing PHI categories:** While overall F1-scores reported by the reviewed studies are quite high, especially at the binary token level, our findings demonstrate that performance of the de-identification systems for some PHI categories (contact information, identifiers, locations, names, and professions) requires further improvement. While contact information and identifiers could most likely be easily improved with rule-based taggers, locations, names and professions remain a challenge. As demonstrated by [67], [83], access to structured data from the EHRs can boost the performance for the *Name* category. We hypothesise that improvements for the *Location* category could be achieved similarly, so integrating EHR data (if available) into the de-identification approach is a research direction that should be further pursued. Additionally, Zhou et al. [81] proposed a method for effective recognition of names based on a set of landmark names and word+contextualised embeddings, a method they claim would be benenficial for locations as well. [18] achieved recall of 99% for patient names, identifiers and phone numbers thanks to supplementing data from other datasets and data augmentation with obscured de-identified data, which allowed improvements of as much as 3 points on some datasets. This type of data augmentation seems like another research direction worth pursuing.

**Fine-tuning and expanding of word and character representation approaches:** We have observed that RNN based methods tend to provide better results for locations and professions than CRF-based systems (Appendix E). Further analysis revealed that these categories are very broadly defined and expressed with high lexical variability that could not be handled by feature engineering methods and dictionaries [42], [61]. However, word embeddings (used by the RNN models) can capture the semantic similarity between lexically different terms [15]. The results of Tang et al. [17] and Liao et al. [37] show that applying (or fine-tuning) state-of-the-art word embedding methods is a promising research direction in this context. Furthermore, Liao et al. [37] found that fine-tuning GloVe embeddings improved domain adaptation performance, unlike simply applying BERT embeddings, a finding confirmed by [23]. Rosario et al. [55] reported that using character-level embeddings improved correct recognition of abbreviated forms or acronyms, morpho-syntactic variations, and typos and helped in making a distinction between polysemous forms. The findings in Ahmed et al. [57] are that char2vec improved the precision for *Location* and *Date* significantly, but at the cost of recall. It also improved the ability to predict labels for OOV tokens.

**Domain transferring:** Our findings show that developing a robust off-the-shelf de-identification system that will perform equally well on documents from different clinical domains and institutions remains an open challenge. Based on our review, we identified different ways of addressing this challenge that represent opportunities for further work:

1. **Annotation ambiguities:** During qualitative evaluation in [19] it was revealed that some cases of false negatives appear due to annotation amgibuities in different corpora, leaving it unclear if drops in performance are due to different labeling standards. Those findings are confirmed in [27]. Having single unified annotation guidelines would undoubtedly alleviate the performance drops during domain transfer.

2. **Annotated data availability:** Ideally, with large amounts of diverse training data generalisability of off-the-shelf systems would be an achievable goal. Currently, training data diversity is a significant challenge. All the publicly available corpora used by the de-identification approaches come from just two different sources, the Partners Healthcare, and Beth Israel Deaconess Medical Center (Table 2). Shared tasks have played an essential role in methodological developments, but relying on shared tasks as a primary source of training data should not be a long-term solution for the clinical NLP community. Synthetic health data that includes health records of realistic (but not real) patients is a viable alternative. Open source synthetic health data-generating tools such as Synthea [94], can increase training data diversity. Annotating synthetic data through crowdsourcing could be an option in case of de-identification, as gold standard entities are not medical concepts. However, [19] point out that synthesized unrealistic dates may weaken generalizations of models to real-life text. An additional way of obtaining diverse training data is to encourage sharing of proprietary corpora, which could be done if further developments in the area of pseudonymised embeddings [10] are pursued.

3. **Utilizing unlabelled data:** Until large amounts of training data become available, research opportunities lie in using unlabelled data or reducing the amount labelling. Unlabelled data has been used to create anonymyzed word embeddings for downstream tasks [11], customise off-the-shelf systems through word embeddings [27] or create pseudo-labels [37]. These directions could be explored further, especially given the recent state-of-the-art contextualised embedding methods [14], [18]–[20], [37], [38], [40], [54]–[58], [81]. Small amounts of training data have been used to boost the performance of machine learning based systems through semi-supervised methods such as active learning [78] and self-training [32], [33], [37]. While active learning requires (at least one) expert annotator, self-training and similar methods (e.g. co-training) could be more extensively tested given that [32], [33], [37] used rather large amounts of straining (seed) training data (e.g., the entire 2006 i2b2 training set). Finally, in cases where training data is available, but not enough to achieve reliable de-identification performance, it can be expanded with publicly available corpora using domain adaptation techniques [37], [63], [88] or even simple merging [52].

**Integration of available approaches:** As our review has demonstrated, self-attention-based models are currently best performing stand-alone models in clinical de-identification. However, several studies have shown that if feature engineering, rule-based methods or even existing off-the-shelf tools are available, they can be integrated with RNNs [15], [44], [53], [62] and transformers [18] to further boost their final F1-score. Ensemble methods of [20], [53] have proven to obtain state-of-the-art results. While methodological efforts in this direction have been made on proprietary corpora [77], developing an open-source extensible de-identification framework that enables integration of different methods and systems is a promising research opportunity.

## 3.7 Limitations

To the best of our knowledge, this is the first systematic review of the clinical free-text NLP de-identification approaches since the work of Meystre et al. in 2010 [12]. Several limitations of our study are discussed below. This study is focused on the clinical domain, so any relevant research in other domains (e.g. legal [95], [96]) has not been included. We do not consider this as a major limitation, as de-identification is a sub-task of NER, and with our study, we aimed to find out about any methodologies and performances that were specific to the clinical domain. Our review is also limited to approaches conducted on English free text. During the screening process we found that de-identification has been studied in Chinese [97]–[100], Japanese [101], Dutch [102], French [103]–[105], Italian [106], [107], German [108]–[110], Spanish [111]–[115], and Swedish [116]–[118]. These studies would provide us with important information, so future work may seek to address them. Finally, as in any systematic review, the information extracted from the studies may be incomplete. Some of the missing information could have been described implicitly or might not be described. It is also possible that we have not managed to find all relevant studies. Our goal was to cover both medical and computer science data sources, so we limited our search to PubMed, Web of Science, and DBLP.

# 4 Conclusion and future work

This review presents and discusses the research in the field of de-identification of clinical free text that has been published since 2010. By reviewing 69 relevant papers, we have found that:

1. Earlier approaches are mainly rule and machine learning hybrids with extensive feature engineering and post-processing, but recent performance improvements are due to feature-inferring RNNs and self-attention-based neural networks.
2. Current state-of-the-art systems provide very high binary token F1-scores (over 98%). However, their performance needs to be more thoroughly assessed with different measures in order to judge their reliability to safely de-identify data in a real-world setting.
3. Without additional manually labelled training data, state-of-the-art systems cannot achieve performance levels needed for safe de-identification of free-text across a wide range of clinical sub-domains.
4. Key challenges in the domain of clinical de-identification include the lack of diverse annotated corpora, and improving the existing or developing new domain adaptation methods.

We have also identified several topics that represent opportunities for further work, including:

1. Developing an industry-wide standard for annotation and performance requirements of de-identification systems.
2. Annotating corpora from a variety of clinical sub-domains, or exploring different ways to create them, such as the generation of synthetic data or distant supervision [119].
3. Finding more efficient ways to use and share the currently available annotated corpora through semi-supervised learning and pseudonymised embeddings.
4. Exploring state-of-the-art word embedding methods pre-trained on general [120], [121], biomedical [122], [123] or clinical domain [124], both in the context of performance improvement and domain adaptation.
5. Exploring other state-of-the-art named entity recognition approaches that have not been used for de-identification, such as zero-shot learning [125], [126] and differentiable neural architecture search [127].

**Acknowledgments**

The authors gratefully acknowledge the support from the Engineering and Physical Sciences Research Council for HealTex—UK Healthcare Text Analytics Research Network (Grant number EP/N027280/1).

**References**

[1] I. Calapodescu, D. Rozier, S. Artemova, and J.-L. Bosson, *Semi-automatic de-identification of hospital discharge summaries with natural language processing A case-study of performance and real-world usability*. New York: Ieee, 2017, pp. 1106–1111. doi: 10.1109/iThings-GreenCom-CPSCom-SmartData.2017.169.

[2] S. Schneeweiss and J. Avorn, "A review of uses of health care utilization databases for epidemiologic research on therapeutics," *J Clin Epidemiol*, vol. 58, no. 4, pp. 323–337, Apr. 2005, doi: 10.1016/j.jclinepi.2004.10.012.

[3] V. Tresp, J. Marc Overhage, M. Bundschus, S. Rabizadeh, P. A. Fasching, and S. Yu, "Going Digital: A Survey on Digitalization and Large-Scale Data Analytics in Healthcare," *Proceedings of the IEEE*, vol. 104, no. 11, pp. 2180–2206, Nov. 2016, doi: 10.1109/JPROC.2016.2615052.

[4] R. N. Cardinal, "Clinical records anonymisation and text extraction (CRATE): an open-source software system," *BMC Med Inform Decis Mak*, vol. 17, no. 1, Art. no. 1, 26 2017, doi: 10.1186/s12911-017-0437-1.

[5] V. Foufi, C. Gaudet-Blavignac, R. Chevrier, and C. Lovis, "De-Identification of Medical Narrative Data," *Stud Health Technol Inform*, vol. 244, pp. 23–27, 2017.

[6] D. Li *et al.*, "A Frequency-based Strategy of Obtaining Sentences from Clinical Data Repository for Crowdsourcing," in *Medinfo 2015: Ehealth-Enabled Health*, I. N. Sarkar, A. Georgiou, and P. M. D. Marques, Eds., Amsterdam: Ios Press, 2015, pp. 1033–1034. doi: 10.3233/978-1-61499-564-7-1033.

[7] D. S. Carrell, D. J. Cronkite, B. A. Malin, J. S. Aberdeen, and L. Hirschman, "Is the Juice Worth the Squeeze? Costs and Benefits of Multiple Human Annotators for Clinical Text De-identification," *Methods Inf Med*, vol. 55, no. 4, Art. no. 4, Aug. 2016, doi: 10.3414/ME15-01-0122.

[8] D. A. Dorr, W. F. Phillips, S. Phansalkar, S. A. Sims, and J. F. Hurdle, "Assessing the difficulty and time cost of de-identification in clinical narratives," *Methods Inf Med*, vol. 45, no. 3, pp. 246–252, 2006.

[9] M. N. Sadat, M. M. A. Aziz, N. Mohammed, S. Pakhomov, H. Liu, and X. Jiang, "A privacy-preserving distributed filtering framework for NLP artifacts," *BMC Med Inform Decis Mak*, vol. 19, no. 1, Art. no. 1, 07 2019, doi: 10.1186/s12911-019-0867-z.

[10] M. Friedrich, A. Köhn, G. Wiedemann, and C. Biemann, "Adversarial Learning of Privacy-Preserving Text Representations for De-Identification of Medical Records," in *Proceedings of the 57th Conference of the Association for Computational Linguistics, ACL 2019, Florence, Italy, July 28- August 2, 2019, Volume 1: Long Papers*, A. Korhonen, D. R. Traum, and L. Màrquez, Eds., Association for Computational Linguistics, 2019, pp. 5829–5839. doi: 10.18653/v1/p19-1584.

[11] A. M, A. M, R. F, and H. G, “Using word embeddings to improve the privacy of clinical notes,” *Journal of the American Medical Informatics Association : JAMIA*, vol. 27, no. 6, pp. 901–907, Jun. 2020, doi: 10.1093/jamia/ocaa038.

[12] S. M. Meystre, F. J. Friedlin, B. R. South, S. Shen, and M. H. Samore, “Automatic de-identification of textual documents in the electronic health record: a review of recent research,” *BMC Med Res Methodol*, vol. 10, p. 70, Aug. 2010, doi: 10.1186/1471-2288-10-70.

[13] A. Stubbs, M. Filannino, and Ö. Uzuner, “De-identification of psychiatric intake records: Overview of 2016 CEGS N-GRID shared tasks Track 1,” *J Biomed Inform*, vol. 75S, pp. S4–S18, Nov. 2017, doi: 10.1016/j.jbi.2017.06.011.

[14] A. Stubbs, C. Kotfila, and Ö. Uzuner, “Automated systems for the de-identification of longitudinal clinical narratives: Overview of 2014 i2b2/UTHealth shared task Track 1,” *J Biomed Inform*, vol. 58 Suppl, pp. S11-19, Dec. 2015, doi: 10.1016/j.jbi.2015.06.007.

[15] F. Dernoncourt, J. Y. Lee, O. Uzuner, and P. Szolovits, “De-identification of patient notes with recurrent neural networks,” *J Am Med Inform Assoc*, vol. 24, no. 3, Art. no. 3, May 2017, doi: 10.1093/jamia/ocw156.

[16] Z. Liu *et al.*, “Automatic de-identification of electronic medical records using token-level and character-level conditional random fields,” *J Biomed Inform*, vol. 58 Suppl, pp. S47-52, Dec. 2015, doi: 10.1016/j.jbi.2015.06.009.

[17] B. Tang, D. Jiang, Q. Chen, X. Wang, J. Yan, and Y. Shen, “De-identification of Clinical Text via Bi-LSTM-CRF with Neural Language Models,” *AMIA Annu Symp Proc*, vol. 2019, pp. 857–863, Mar. 2020.

[18] C. PJ, W. C, S. JM, A. J, C. TS, and L. CP, “Automated deidentification of radiology reports combining transformer and ‘hide in plain sight’ rule-based methods.,” *Journal of the American Medical Informatics Association : JAMIA*, vol. 30, no. 2, pp. 318–328, Jan. 2023, doi: 10.1093/jamia/ocac219.

[19] A. E. W. Johnson, L. Bulgarelli, and T. J. Pollard, “Deidentification of free-text medical records using pre-trained bidirectional transformers,” in *ACM CHIL ’20: ACM Conference on Health, Inference, and Learning, Toronto, Ontario, Canada, April 2-4, 2020 [delayed]*, M. Ghassemi, Ed., ACM, 2020, pp. 214–221. doi: 10.1145/3368555.3384455.

[20] L. L, P.-C. O, N. A, B. V, and J. L, “De-identifying Australian hospital discharge summaries: An end-to-end framework using ensemble of deep learning models,” *Journal of biomedical informatics*, vol. 135, p. 104215, Nov. 2022, doi: 10.1016/j.jbi.2022.104215.

[21] B. Kitchenham, “Procedures for Performing Systematic Reviews,” *Keele, UK, Keele Univ.*, vol. 33, Aug. 2004.

[22] *Cochrane Handbook for Systematic Reviews of Interventions*, 1 edition. Chichester, England ; Hoboken, NJ: Wiley, 2008.

[23] O. Uzuner, Y. Luo, and P. Szolovits, "Evaluating the State-of-the-Art in Automatic De-identification," *Journal of the American Medical Informatics Association*, vol. 14, no. 5, pp. 550–563, Sep. 2007, doi: 10.1197/jamia.M2444.

[24] J. Aberdeen *et al.*, "The MITRE Identification Scrubber Toolkit: Design, training, and assessment," *Int. J. Med. Inform.*, vol. 79, no. 12, Art. no. 12, Dec. 2010, doi: 10.1016/j.ijmedinf.2010.09.007.

[25] L. Deleger *et al.*, "Large-scale evaluation of automated clinical note de-identification and its impact on information extraction," *J Am Med Inform Assoc*, vol. 20, no. 1, Art. no. 1, Jan. 2013, doi: 10.1136/amiajnl-2012-001012.

[26] J. J. Gardner, L. Xiong, F. Wang, A. R. Post, J. H. Saltz, and T. Grandison, "An evaluation of feature sets and sampling techniques for de-identification of medical records," in *ACM International Health Informatics Symposium, IHI 2010, Arlington, VA, USA, November 11 - 12, 2010, Proceedings*, T. C. Veinot, Ü. V. Çatalyürek, G. Luo, H. Andrade, and N. R. Smalheiser, Eds., ACM, 2010, pp. 183–190. doi: 10.1145/1882992.1883019.

[27] T. Hartman *et al.*, "Customization scenarios for de-identification of clinical notes," *BMC Med Inform Decis Mak*, vol. 20, no. 1, Art. no. 1, Jan. 2020, doi: 10.1186/s12911-020-1026-2.

[28] K. Li, Y. Chai, H. Zhao, X. Nan, and Y. Zhao, "Learning to Recognize Protected Health Information in Electronic Health Records with Recurrent Neural Network," in *Natural Language Understanding and Intelligent Applications (nlpcc 2016)*, C. Y. Lin, N. Xue, D. Zhao, X. Huang, and Y. Feng, Eds., Cham: Springer International Publishing Ag, 2016, pp. 575–582. doi: 10.1007/978-3-319-50496-4_51.

[29] M. Li, D. Carrell, J. Aberdeen, L. Hirschman, and B. A. Malin, "De-identification of clinical narratives through writing complexity measures," *Int J Med Inform*, vol. 83, no. 10, Art. no. 10, Oct. 2014, doi: 10.1016/j.ijmedinf.2014.07.002.

[30] X.-B. Li and J. Qin, "Anonymizing and Sharing Medical Text Records," *Inf Syst Res*, vol. 28, no. 2, Art. no. 2, 2017, doi: 10.1287/isre.2016.0676.

[31] A. J. McMurry, B. Fitch, G. Savova, I. S. Kohane, and B. Y. Reis, "Improved de-identification of physician notes through integrative modeling of both public and private medical text," *BMC Med Inform Decis Mak*, vol. 13, p. 112, Oct. 2013, doi: 10.1186/1472-6947-13-112.

[32] N. D. Phuong and V. T. N. Chau, "Automatic de-identification of medical records with a multilevel hybrid semi-supervised learning approach," in *2016 IEEE RIVF International Conference on Computing & Communication Technologies, Research, Innovation, and Vision for the Future, RIVF 2016, Hanoi, Vietnam, November 7-9, 2016*, T. Cao and Y.-S. Ho, Eds., IEEE, 2016, pp. 43–48. doi: 10.1109/RIVF.2016.7800267.

[33] N. D. Phuong, V. T. N. Chau, and H. T. Bao, "A Hybrid Semi-supervised Learning Approach to Identifying Protected Health Information in Electronic Medical Records," in *Proceedings of the 10th International Conference on Ubiquitous Information Management and Communication, IMCOM 2016, Danang, Vietnam, January 4-6, 2016*, ACM, 2016, p. 82:1-82:8. doi: 10.1145/2857546.2857630.

[34] Y.-S. Zhao, K.-L. Zhang, H.-C. Ma, and K. Li, "Leveraging text skeleton for de-identification of electronic medical records," *BMC Med Inform Decis Mak*, vol. 18, no. Suppl 1, Art. no. Suppl 1, 22 2018, doi: 10.1186/s12911-018-0598-6.

[35] G. Zuccon, D. Kotzur, A. Nguyen, and A. Bergheim, "De-identification of health records using Anonym: effectiveness and robustness across datasets," *Artif Intell Med*, vol. 61, no. 3, Art. no. 3, Jul. 2014, doi: 10.1016/j.artmed.2014.03.006.

[36] O. Ferrández, B. R. South, S. Shen, F. J. Friedlin, M. H. Samore, and S. M. Meystre, "BoB, a best-of-breed automated text de-identification system for VHA clinical documents," *J Am Med Inform Assoc*, vol. 20, no. 1, pp. 77–83, Jan. 2013, doi: 10.1136/amiajnl-2012-001020.

[37] S. Liao, J. Kiros, J. Chen, Z. Zhang, and T. Chen, "Improving domain adaptation in de-identification of electronic health records through self-training," *J. Am. Medical Informatics Assoc.*, vol. 28, no. 10, pp. 2093–2100, 2021, doi: 10.1093/jamia/ocab128.

[38] L. K, K. M, H. S, and U. Ö, "A Context-Enhanced De-identification System," *ACM transactions on computing for healthcare*, vol. 3, no. 1, Jan. 2022, doi: 10.1145/3470980.

[39] A. Stubbs and Ö. Uzuner, "Annotating longitudinal clinical narratives for de-identification: The 2014 i2b2/UTHealth corpus," *J Biomed Inform*, vol. 58 Suppl, pp. S20-29, Dec. 2015, doi: 10.1016/j.jbi.2015.07.020.

[40] A. T, A. MMA, and M. N, "De-identification of electronic health record using neural network," *Scientific reports*, vol. 10, no. 1, p. 18600, Oct. 2020, doi: 10.1038/s41598-020-75544-1.

[41] T. Chen, R. M. Cullen, and M. Godwin, "Hidden Markov model using Dirichlet process for de-identification," *J Biomed Inform*, vol. 58 Suppl, pp. S60-66, Dec. 2015, doi: 10.1016/j.jbi.2015.09.004.

[42] A. Dehghan, A. Kovacevic, G. Karystianis, J. A. Keane, and G. Nenadic, "Combining knowledge- and data-driven methods for de-identification of clinical narratives," *J Biomed Inform*, vol. 58 Suppl, pp. S53-59, Dec. 2015, doi: 10.1016/j.jbi.2015.06.029.

[43] B. He, Y. Guan, J. Cheng, K. Cen, and W. Hua, "CRFs based de-identification of medical records," *J Biomed Inform*, vol. 58 Suppl, pp. S39-46, Dec. 2015, doi: 10.1016/j.jbi.2015.08.012.

[44] Y. Kim, P. Heider, and S. Meystre, “Ensemble-based Methods to Improve De-identification of Electronic Health Record Narratives,” *AMIA Annu Symp Proc*, vol. 2018, pp. 663–672, 2018.

[45] K. Lee, M. Filannino, and Ö. Uzuner, “An Empirical Test of GRUs and Deep Contextualized Word Representations on De-Identification,” *Stud Health Technol Inform*, vol. 264, pp. 218–222, Aug. 2019, doi: 10.3233/SHTI190215.

[46] Z. Liu, B. Tang, X. Wang, and Q. Chen, “De-identification of clinical notes via recurrent neural network and conditional random field,” *J Biomed Inform*, vol. 75S, pp. S34–S42, Nov. 2017, doi: 10.1016/j.jbi.2017.05.023.

[47] Z. Liu *et al.*, “Entity recognition from clinical texts via recurrent neural network,” *BMC Med Inform Decis Mak*, vol. 17, no. Suppl 2, Art. no. Suppl 2, Jul. 2017, doi: 10.1186/s12911-017-0468-7.

[48] A. Madan, A. M. George, A. Singh, and M. P. S. Bhatia, “Redaction of Protected Health Information in EHRs using CRFs and Bi-directional LSTMs,” in *2018 7th International Conference on Reliability, Infocom Technologies and Optimization (trends and Future Directions) (icrito) (icrito)*, New York: Ieee, 2018, pp. 513–517.

[49] Shweta, A. Kumar, A. Ekbal, S. Saha, and P. Bhattacharyya, “A Recurrent Neural Network Architecture for De-identifying Clinical Records,” in *Proceedings of the 13th International Conference on Natural Language Processing, ICON 2016, Varanasi, India, December 17-20, 2016*, D. M. Sharma, R. Sangal, and A. K. Singh, Eds., NLP Association of India, 2016, pp. 188–197. [Online]. Available: https://www.aclweb.org/anthology/W16-6325/

[50] S. Yadav, A. Ekbal, S. Saha, and P. Bhattacharyya, “Deep Learning Architecture for Patient Data De-identification in Clinical Records,” in *Proceedings of the Clinical Natural Language Processing Workshop, ClinicalNLP@COLING 2016, Osaka, Japan, December 11, 2016*, A. Rumshisky, K. Roberts, S. Bethard, and T. Naumann, Eds., The COLING 2016 Organizing Committee, 2016, pp. 32–41. [Online]. Available: https://www.aclweb.org/anthology/W16-4206/

[51] H. Yang and J. M. Garibaldi, “Automatic detection of protected health information from clinic narratives,” *J Biomed Inform*, vol. 58 Suppl, pp. S30-38, Dec. 2015, doi: 10.1016/j.jbi.2015.06.015.

[52] X. Yang, T. Lyu, C.-Y. Lee, J. Bian, W. R. Hogan, and Y. Wu, “A Study of Deep Learning Methods for De-identification of Clinical Notes at Cross Institute Settings,” *IEEE Int Conf Healthc Inform*, vol. 2019, Jun. 2019, doi: 10.1109/ICHI.2019.8904544.

[53] K. Y, H. PM, and M. SM, “Comparative Study of Various Approaches for Ensemble-based De-identification of Electronic Health Record Narratives,” in *AMIA ... Annual Symposium proceedings. AMIA Symposium*, United States: Medical University of South Carolina, Charleston, South Carolina, USA.; Medical University of South Carolina, Charleston, South Carolina, USA.; Medical University of South Carolina, Charleston, South

Carolina, USA.; Clinacuity, Inc., Charleston, South Carolina, USA., 2020, pp. 648–657. [Online]. Available: https://pubmed.ncbi.nlm.nih.gov/33936439/

[54] M. Abadeer, "Assessment of DistilBERT performance on Named Entity Recognition task for the detection of Protected Health Information and medical concepts," in *Proceedings of the 3rd Clinical Natural Language Processing Workshop, ClinicalNLP@EMNLP 2020, Online, November 19, 2020*, A. Rumshisky, K. Roberts, S. Bethard, and T. Naumann, Eds., Association for Computational Linguistics, 2020, pp. 158–167. doi: 10.18653/v1/2020.clinicalnlp-1.18.

[55] R. Catelli, V. Casola, G. De Pietro, H. Fujita, and M. Esposito, "Combining contextualized word representation and sub-document level analysis through Bi-LSTM+CRF architecture for clinical de-identification," *Knowledge-Based Systems*, vol. 213, p. 106649, Feb. 2021, doi: 10.1016/j.knosys.2020.106649.

[56] Md. M. Anjum, N. Mohammed, and X. Jiang, "De-identification of Unstructured Clinical Texts from Sequence to Sequence Perspective," in *CCS '21: 2021 ACM SIGSAC Conference on Computer and Communications Security*, Y. Kim, J. Kim, G. Vigna, and E. Shi, Eds., Republic of Korea: ACM, 2021, pp. 2438–2440. doi: 10.1145/3460120.3485354.

[57] A. A, A. A, and E. C, "Benchmarking Modern Named Entity Recognition Techniques for Free-text Health Record Deidentification.," in *AMIA Joint Summits on Translational Science proceedings*, United States: Brown University, Providence, RI, United States.; Brown University, Providence, RI, United States.; Brown University, Providence, RI, United States., 2021, pp. 102–111. [Online]. Available: https://pubmed.ncbi.nlm.nih.gov/34457124/

[58] M. K *et al.*, "Building a best-in-class automated de-identification tool for electronic health records through ensemble learning.," *Patterns (New York, N.Y.)*, vol. 2, no. 6, p. 100255, Jun. 2021, doi: 10.1016/j.patter.2021.100255.

[59] F. ZN, C.-D. A, U. LH, and W. GE, "Word embeddings trained on published case reports are lightweight, effective for clinical tasks, and free of protected health information.," *Journal of biomedical informatics*, vol. 125, p. 103971, Jan. 2022, doi: 10.1016/j.jbi.2021.103971.

[60] D. D. A. Bui, M. Wyatt, and J. J. Cimino, "The UAB Informatics Institute and 2016 CEGS N-GRID de-identification shared task challenge," *J Biomed Inform*, vol. 75S, pp. S54–S61, Nov. 2017, doi: 10.1016/j.jbi.2017.05.001.

[61] A. Dehghan, A. Kovacevic, G. Karystianis, J. A. Keane, and G. Nenadic, "Learning to identify Protected Health Information by integrating knowledge- and data-driven algorithms: A case study on psychiatric evaluation notes," *J Biomed Inform*, vol. 75S, pp. S28–S33, Nov. 2017, doi: 10.1016/j.jbi.2017.06.005.

[62] Z. Jiang, C. Zhao, B. He, Y. Guan, and J. Jiang, "De-identification of medical records using conditional random fields and long short-term memory networks," *J Biomed Inform*, vol. 75S, pp. S43–S53, Nov. 2017, doi: 10.1016/j.jbi.2017.10.003.

[63] H.-J. Lee, Y. Zhang, K. Roberts, and H. Xu, "Leveraging existing corpora for de-identification of psychiatric notes using domain adaptation," *AMIA Annu Symp Proc*, vol. 2017, pp. 1070–1079, 2017.

[64] Joon Lee, D. J. Scott, M. Villarroel, G. D. Clifford, M. Saeed, and R. G. Mark, "Open-access MIMIC-II database for intensive care research," in *2011 Annual International Conference of the IEEE Engineering in Medicine and Biology Society*, Boston, MA: IEEE, Aug. 2011, pp. 8315–8318. doi: 10.1109/IEMBS.2011.6092050.

[65] A. L. Goldberger *et al.*, "PhysioBank, PhysioToolkit, and PhysioNet: components of a new research resource for complex physiologic signals," *Circulation*, vol. 101, no. 23, pp. E215-220, Jun. 2000, doi: 10.1161/01.cir.101.23.e215.

[66] I. Neamatullah *et al.*, "Automated de-identification of free-text medical records," *BMC Med Inform Decis Mak*, vol. 8, no. 1, p. 32, Dec. 2008, doi: 10.1186/1472-6947-8-32.

[67] J. Y. Lee, F. Dernoncourt, Ö. Uzuner, and P. Szolovits, "Feature-Augmented Neural Networks for Patient Note De-identification," in *Proceedings of the Clinical Natural Language Processing Workshop, ClinicalNLP@COLING 2016, Osaka, Japan, December 11, 2016*, A. Rumshisky, K. Roberts, S. Bethard, and T. Naumann, Eds., The COLING 2016 Organizing Committee, 2016, pp. 17–22. [Online]. Available: https://www.aclweb.org/anthology/W16-4204/

[68] S. Gehrmann *et al.*, "Comparing deep learning and concept extraction based methods for patient phenotyping from clinical narratives," *PLoS ONE*, vol. 13, no. 2, p. e0192360, Feb. 2018, doi: 10.1371/journal.pone.0192360.

[69] Ó. Ferrández, B. R. South, S. Shen, and S. M. Meystre, "A Hybrid Stepwise Approach for De-identifying Person Names in Clinical Documents," in *Proceedings of the 2012 Workshop on Biomedical Natural Language Processing, BioNLP@HLT-NAACL Montrèal, Canada, June 8, 2012*, K. B. Cohen, D. Demner-Fushman, S. Ananiadou, B. L. Webber, J. Tsujii, and J. Pestian, Eds., Association for Computational Linguistics, 2012, pp. 65–72. [Online]. Available: https://www.aclweb.org/anthology/W12-2408/

[70] D. Hanauer *et al.*, "Bootstrapping a de-identification system for narrative patient records: cost-performance tradeoffs," *Int J Med Inform*, vol. 82, no. 9, Art. no. 9, Sep. 2013, doi: 10.1016/j.ijmedinf.2013.03.005.

[71] A. C. Fernandes *et al.*, "Development and evaluation of a de-identification procedure for a case register sourced from mental health electronic records," *BMC Med Inform Decis Mak*, vol. 13, p. 71, Jul. 2013, doi: 10.1186/1472-6947-13-71.

[72] M. Kayaalp, A. C. Browne, Z. A. Dodd, P. Sagan, and C. J. McDonald, "De-identification of Address, Date, and Alphanumeric Identifiers in Narrative Clinical Reports," in *AMIA 2014, American Medical Informatics Association Annual Symposium, Washington, DC, USA, November 15-19, 2014*, AMIA, 2014. [Online]. Available:

http://knowledge.amia.org/56638-amia-1.1540970/t-004-1.1544972/f-004-1.1544973/a-174-1.1545151/a-175-1.1545148

[73] A. Redd *et al.*, "Evaluation of PHI Hunter in Natural Language Processing Research," *Perspect Health Inf Manag*, vol. 12, p. 1f, 2015.

[74] J. Seeger, A. Culotta, J. Keller, P. van Kessel, and M. Jugovich, *Data Deidentification in Medical Transcriptions using Regular Expressions and Machine Learning*. New York: Ieee, 2015, pp. 1322–1329.

[75] A. Dehghan, T. Liptrot, D. Tibble, M. Barker-Hewitt, and G. Nenadic, "Identification of Occupation Mentions in Clinical Narratives," in *Natural Language Processing and Information Systems - 21st International Conference on Applications of Natural Language to Information Systems, NLDB 2016, Salford, UK, June 22-24, 2016, Proceedings*, E. Métais, F. Meziane, M. Saraee, V. Sugumaran, and S. Vadera, Eds., in Lecture Notes in Computer Science, vol. 9612. Springer, 2016, pp. 359–365. doi: 10.1007/978-3-319-41754-7_35.

[76] S. Polsley, A. Tahir, M. Raju, A. Akinleye, and D. Steward, "Role-Preserving Redaction of Medical Records to Enable Ontology-Driven Processing," in *BioNLP 2017, Vancouver, Canada, August 4, 2017*, K. B. Cohen, D. Demner-Fushman, S. Ananiadou, and J. Tsujii, Eds., Association for Computational Linguistics, 2017, pp. 194–199. doi: 10.18653/v1/W17-2324.

[77] S. Braghin, J. H. Bettencourt-Silva, K. Levacher, and S. Antonatos, "An Extensible De-Identification Framework for Privacy Protection of Unstructured Health Information: Creating Sustainable Privacy Infrastructures," *Stud Health Technol Inform*, vol. 264, pp. 1140–1144, Aug. 2019, doi: 10.3233/SHTI190404.

[78] M. Li, M. Scaiano, K. El Emam, and B. A. Malin, "Efficient Active Learning for Electronic Medical Record De-identification," *AMIA Jt Summits Transl Sci Proc*, vol. 2019, pp. 462–471, 2019.

[79] S. JM, P. T, A. J, K. C. Jr, and C. TS, "Evaluation of Automated Public De-Identification Tools on a Corpus of Radiology Reports.," *Radiology. Artificial intelligence*, vol. 2, no. 6, p. e190137, Nov. 2020, doi: 10.1148/ryai.2020190137.

[80] A. E. W. Johnson *et al.*, "MIMIC-III, a freely accessible critical care database," *Sci Data*, vol. 3, no. 1, p. 160035, May 2016, doi: 10.1038/sdata.2016.35.

[81] Z. H and R. D, "An embedding-based medical note de-identification approach with sparse annotation.," *Medical physics*, vol. 48, no. 3, pp. 1341–1348, Mar. 2021, doi: 10.1002/mp.14664.

[82] T. Paul *et al.*, "Utility of Features in a Natural-Language-Processing-Based Clinical De-Identification Model Using Radiology Reports for Advanced NSCLC Patients," *APPLIED SCIENCES-BASEL*, vol. 12, no. 19, Oct. 2022, doi: 10.3390/app12199976.

[83] A.-E.-R. N *et al.*, “Natural Language Processing for Enterprise-scale De-identification of Protected Health Information in Clinical Notes.,” in *AMIA ... Annual Symposium proceedings. AMIA Symposium*, United States: Division of Medical Informatics, Department of Internal Medicine, University of Kansas Medical Center, Kansas City, Kansas.; Medical College of Wisconsin, Milwaukee, Wisconsin.; Medical College of Wisconsin, Milwaukee, Wisconsin.; Medical College of Wisconsin, Milwaukee, Wisconsin.; The University of Kansas Hospital, Kansas City, Kansas.; Division of Medical Informatics, Department of Internal Medicine, University of Kansas Medical Center, Kansas City, Kansas.; Medical College of Wisconsin, Milwaukee, Wisconsin.; Division of Health Management and Informatics, University of Missouri, Columbia, Missouri., 2022, pp. 92–101. [Online]. Available: https://pubmed.ncbi.nlm.nih.gov/35854742/

[84] S. M *et al.*, “DeIDNER Model: A Neural Network Named Entity Recognition Model for Use in the De-identification of Clinical Notes.,” in *Biomedical engineering systems and technologies, international joint conference, BIOSTEC*, Germany: Department of Biomedical Informatics, University of Arkansas for Medical Sciences, Little Rock, AR, U.S.A.; College of Medicine, University of Arkansas for Medical Sciences, Little Rock, AR, U.S.A.; Department of Biological Sciences and Arkansas Biosciences Institute, Arkansas State University, Jonesboro, AR, U.S.A.; Department of Population Health Sciences, University of Texas Health Science Center at San Antonio, San Antonio, TX, U.S.A., Feb. 2022, pp. 640–647. doi: 10.5220/0010884500003123.

[85] P. T *et al.*, “Investigation of the Utility of Features in a Clinical De-identification Model: A Demonstration Using EHR Pathology Reports for Advanced NSCLC Patients.,” *Frontiers in digital health*, vol. 4, p. 728922, 2022, doi: 10.3389/fdgth.2022.728922.

[86] K. DP, S. RI, F. M, N. YN, B. K, and V. CM, “Evaluation of an automated Presidio anonymisation model for unstructured radiation oncology electronic medical records in an Australian setting.,” *International journal of medical informatics*, vol. 168, p. 104880, Dec. 2022, doi: 10.1016/j.ijmedinf.2022.104880.

[87] S. Jain *et al.*, “RadGraph: Extracting Clinical Entities and Relations from Radiology Reports.” arXiv, Aug. 29, 2021. Accessed: Apr. 06, 2023. [Online]. Available: http://arxiv.org/abs/2106.14463

[88] H.-J. Lee, Y. Wu, Y. Zhang, J. Xu, H. Xu, and K. Roberts, “A hybrid approach to automatic de-identification of psychiatric notes,” *J Biomed Inform*, vol. 75S, pp. S19–S27, Nov. 2017, doi: 10.1016/j.jbi.2017.06.006.

[89] M. Kayaalp *et al.*, “The pattern of name tokens in narrative clinical text and a comparison of five systems for redacting them,” *J Am Med Inform Assoc*, vol. 21, no. 3, Art. no. 3, Jun. 2014, doi: 10.1136/amiajnl-2013-001689.

[90] J. D. Lafferty, A. McCallum, and F. C. N. Pereira, “Conditional Random Fields: Probabilistic Models for Segmenting and Labeling Sequence Data,” in *Proceedings of the*

*Eighteenth International Conference on Machine Learning*, in ICML '01. San Francisco, CA, USA: Morgan Kaufmann Publishers Inc., Jun. 2001, pp. 282–289.

[91] D. D. A. Bui, D. T. Redden, and J. J. Cimino, "Is Multiclass Automatic Text De-Identification Worth the Effort?," *Methods Inf Med*, vol. 57, no. 4, Art. no. 4, 2018, doi: 10.3414/ME18-01-0017.

[92] X. Yang *et al.*, "A study of deep learning methods for de-identification of clinical notes in cross-institute settings," *BMC Med Inform Decis Mak*, vol. 19, no. Suppl 5, Art. no. Suppl 5, 05 2019, doi: 10.1186/s12911-019-0935-4.

[93] Y. Ganin *et al.*, "Domain-adversarial training of neural networks," *J. Mach. Learn. Res.*, vol. 17, no. 1, pp. 2096–2030, Jan. 2016.

[94] J. Walonoski *et al.*, "Synthea: An approach, method, and software mechanism for generating synthetic patients and the synthetic electronic health care record," *Journal of the American Medical Informatics Association*, vol. 25, no. 3, pp. 230–238, Mar. 2018, doi: 10.1093/jamia/ocx079.

[95] V. Barriere and A. Fouret, "May I Check Again? -- A simple but efficient way to generate and use contextual dictionaries for Named Entity Recognition. Application to French Legal Texts," *arXiv:1909.03453 [cs]*, Sep. 2019, Accessed: May 08, 2020. [Online]. Available: http://arxiv.org/abs/1909.03453

[96] D. Garat and D. Wonsever, "Towards De-identification of Legal Texts," *arXiv:1910.03739 [cs]*, Oct. 2019, Accessed: May 08, 2020. [Online]. Available: http://arxiv.org/abs/1910.03739

[97] L. Du, C. Xia, Z. Deng, G. Lu, S. Xia, and J. Ma, "A machine learning based approach to identify protected health information in Chinese clinical text," *Int J Med Inform*, vol. 116, pp. 24–32, 2018, doi: 10.1016/j.ijmedinf.2018.05.010.

[98] Z. Jian *et al.*, "A cascaded approach for Chinese clinical text de-identification with less annotation effort," *J Biomed Inform*, vol. 73, pp. 76–83, 2017, doi: 10.1016/j.jbi.2017.07.017.

[99] Y.-Q. Lee *et al.*, "Protected Health Information Recognition of Unstructured Code-Mixed Electronic Health Records in Taiwan," in *Studies in Health Technology and Informatics*, P. Otero, P. Scott, S. Z. Martin, and E. Huesing, Eds., IOS Press, 2022. doi: 10.3233/SHTI220153.

[100] P. Wang *et al.*, "An Efficient Method for Deidentifying Protected Health Information in Chinese Electronic Health Records: Algorithm Development and Validation," *JMIR Med Inform*, vol. 10, no. 8, p. e38154, Aug. 2022, doi: 10.2196/38154.

[101] K. Kajiyama, H. Horiguchi, T. Okumura, M. Morita, and Y. Kano, "De-identifying free text of Japanese electronic health records," *J Biomed Semant*, vol. 11, no. 1, p. 11, Dec. 2020, doi: 10.1186/s13326-020-00227-9.

[102] V. Menger, F. Scheepers, L. M. van Wijk, and M. Spruit, “DEDUCE: A pattern matching method for automatic de-identification of Dutch medical text,” *Telematics Informatics*, vol. 35, no. 4, Art. no. 4, 2018, doi: 10.1016/j.tele.2017.08.002.

[103] E. Chazard, C. Mouret, G. Ficheur, A. Schaffar, J.-B. Beuscart, and R. Beuscart, “Proposal and evaluation of FASDIM, a Fast And Simple De-Identification Method for unstructured free-text clinical records,” *Int J Med Inform*, vol. 83, no. 4, Art. no. 4, Apr. 2014, doi: 10.1016/j.ijmedinf.2013.11.005.

[104] C. Grouin and P. Zweigenbaum, “Automatic de-identification of French clinical records: comparison of rule-based and machine-learning approaches,” *Stud Health Technol Inform*, vol. 192, pp. 476–480, 2013.

[105] L. Bourdois *et al.*, “De-identification of Emergency Medical Records in French: Survey and Comparison of State-of-the-Art Automated Systems,” *FLAIRS*, vol. 34, no. 1, Apr. 2021, doi: 10.32473/flairs.v34i1.128480.

[106] R. Catelli, F. Gargiulo, V. Casola, G. De Pietro, H. Fujita, and M. Esposito, “Crosslingual named entity recognition for clinical de-identification applied to a COVID-19 Italian data set,” *Applied Soft Computing*, vol. 97, p. 106779, Dec. 2020, doi: 10.1016/j.asoc.2020.106779.

[107] R. Catelli, F. Gargiulo, V. Casola, G. De Pietro, H. Fujita, and M. Esposito, “A Novel COVID-19 Data Set and an Effective Deep Learning Approach for the De-Identification of Italian Medical Records,” *IEEE Access*, vol. 9, pp. 19097–19110, 2021, doi: 10.1109/ACCESS.2021.3054479.

[108] T. Kolditz *et al.*, “Annotating German Clinical Documents for De-Identification,” *Stud Health Technol Inform*, vol. 264, pp. 203–207, Aug. 2019, doi: 10.3233/SHTI190212.

[109] P. Richter-Pechanski, S. Riezler, and C. Dieterich, “De-Identification of German Medical Admission Notes,” *Stud Health Technol Inform*, vol. 253, pp. 165–169, 2018.

[110] M. Baumgartner *et al.*, “Impact Analysis of De-Identification in Clinical Notes Classification,” in *Studies in Health Technology and Informatics*, G. Schreier, B. Pfeifer, M. Baumgartner, and D. Hayn, Eds., IOS Press, 2022. doi: 10.3233/SHTI220368.

[111] F. Hassan, M. Jabreel, N. Maaroof, D. Sánchez, J. Domingo-Ferrer, and A. Moreno, “ReCRF: Spanish Medical Document Anonymization using Automatically-crafted Rules and CRF,” in *Proceedings of the Iberian Languages Evaluation Forum co-located with 35th Conference of the Spanish Society for Natural Language Processing, IberLEF@SEPLN 2019, Bilbao, Spain, September 24th, 2019*, M. Á. G. Cumbreras, J. Gonzalo, E. M. Cámara, R. Martínez-Unanue, P. Rosso, J. Carrillo-de-Albornoz, S. Montalvo, L. Chiruzzo, S. Collovini, Y. Gutiérrez, S. M. J. Zafra, M. Krallinger, M. Montes-y-Gómez, R. Ortega-Bueno, and A. Rosá, Eds., in CEUR Workshop Proceedings, vol. 2421. CEUR-WS.org, 2019, pp. 727–734. [Online]. Available: http://ceur-ws.org/Vol-2421/MEDDOCAN_paper_12.pdf

[112] J. P. Zamorano, “Spanish Medical Document Anonymization with Three-channel Convolutional Neural Networks,” in *Proceedings of the Iberian Languages Evaluation Forum co-located with 35th Conference of the Spanish Society for Natural Language Processing, IberLEF@SEPLN 2019, Bilbao, Spain, September 24th, 2019*, M. Á. G. Cumbreras, J. Gonzalo, E. M. Cámara, R. Martínez-Unanue, P. Rosso, J. Carrillo-de-Albornoz, S. Montalvo, L. Chiruzzo, S. Collovini, Y. Gutiérrez, S. M. J. Zafra, M. Krallinger, M. Montes-y-Gómez, R. Ortega-Bueno, and A. Rosá, Eds., in CEUR Workshop Proceedings, vol. 2421. CEUR-WS.org, 2019, pp. 639–646. [Online]. Available: http://ceur-ws.org/Vol-2421/MEDDOCAN_paper_1.pdf

[113] I. Pérez-Díez, R. Pérez-Moraga, A. López-Cerdán, J.-M. Salinas-Serrano, and M. D. La Iglesia-Vayá, “De-identifying Spanish medical texts - named entity recognition applied to radiology reports,” *J Biomed Semant*, vol. 12, no. 1, p. 6, Dec. 2021, doi: 10.1186/s13326-021-00236-2.

[114] Lima Lopez, Salvador, Perez, Naiara, García-Sardiña, Laura, and Cuadros, Montse, “HitzalMed: Anonymisation of Clinical Text in Spanish,” in *Proceedings of the Twelfth Language Resources and Evaluation Conference*, Marseille, France: European Language Resources Association, 2020, pp. 7038–7043.

[115] García Pablos, Aitor, Perez, Naiara, and Cuadros, Montse, “Sensitive Data Detection and Classification in Spanish Clinical Text: Experiments with BERT,” in *Proceedings of the Twelfth Language Resources and Evaluation Conference*, Marseille, France: European Language Resources Association, 2020, pp. 4486–4494.

[116] H. Dalianis and S. Velupillai, “De-identifying Swedish clinical text - refinement of a gold standard and experiments with Conditional random fields,” *J. Biomedical Semantics*, vol. 1, p. 6, 2010, doi: 10.1186/2041-1480-1-6.

[117] Taridzo Chomutare, Kassaye Yitbarek Yigzaw, Andrius Budrionis, Alexandra Makhlysheva, Fred Godtliebsen, and Hercules Dalianis, “De-Identifying Swedish EHR Text Using Public Resources in the General Domain,” *Studies in health technology and informatics*, pp. 148–152, 2020, doi: 10.3233/shti200140.

[118] Berg, Hanna and Dalianis, Hercules, “A Semi-supervised Approach for De-identification of Swedish Clinical Text,” in *Proceedings of the Twelfth Language Resources and Evaluation Conference*, Marseille, France: European Language Resources Association, 2020, pp. 4444–4450.

[119] M. Mintz, S. Bills, R. Snow, and D. Jurafsky, “Distant supervision for relation extraction without labeled data,” in *Proceedings of the Joint Conference of the 47th Annual Meeting of the ACL and the 4th International Joint Conference on Natural Language Processing of the AFNLP*, Suntec, Singapore: Association for Computational Linguistics, Aug. 2009, pp. 1003–1011. Accessed: May 09, 2020. [Online]. Available: https://www.aclweb.org/anthology/P09-1113

[120] A. Akbik, T. Bergmann, D. Blythe, K. Rasul, S. Schweter, and R. Vollgraf, “FLAIR: An Easy-to-Use Framework for State-of-the-Art NLP,” in *Proceedings of the 2019 Conference of the North American Chapter of the Association for Computational Linguistics (Demonstrations)*, Minneapolis, Minnesota: Association for Computational Linguistics, Jun. 2019, pp. 54–59. doi: 10.18653/v1/N19-4010.

[121] Y. Liu *et al.*, “RoBERTa: A Robustly Optimized BERT Pretraining Approach,” *arXiv:1907.11692 [cs]*, Jul. 2019, Accessed: May 10, 2020. [Online]. Available: http://arxiv.org/abs/1907.11692

[122] J. Lee *et al.*, “BioBERT: a pre-trained biomedical language representation model for biomedical text mining,” *Bioinformatics*, vol. 36, no. 4, pp. 1234–1240, Feb. 2020, doi: 10.1093/bioinformatics/btz682.

[123] Y. Gu *et al.*, “Domain-Specific Language Model Pretraining for Biomedical Natural Language Processing,” *ACM Trans. Comput. Healthcare*, vol. 3, no. 1, pp. 1–23, Jan. 2022, doi: 10.1145/3458754.

[124] E. Alsentzer *et al.*, “Publicly Available Clinical BERT Embeddings,” in *Proceedings of the 2nd Clinical Natural Language Processing Workshop*, Minneapolis, Minnesota, USA: Association for Computational Linguistics, Jun. 2019, pp. 72–78. doi: 10.18653/v1/W19-1909.

[125] M. Košprdić, N. Prodanović, A. Ljajić, B. Bašaragin, and N. Milošević, “From Zero to Hero: Harnessing Transformers for Biomedical Named Entity Recognition in Zero- and Few-shot Contexts.” arXiv, May 27, 2023. Accessed: Jun. 05, 2023. [Online]. Available: http://arxiv.org/abs/2305.04928

[126] Y. Hu *et al.*, “Zero-shot Clinical Entity Recognition using ChatGPT.” Mar. 28, 2023. doi: 10.48550/arXiv.2303.16416.

[127] Y. Jiang, C. Hu, T. Xiao, C. Zhang, and J. Zhu, “Improved Differentiable Architecture Search for Language Modeling and Named Entity Recognition,” in *Proceedings of the 2019 Conference on Empirical Methods in Natural Language Processing and the 9th International Joint Conference on Natural Language Processing (EMNLP-IJCNLP)*, Hong Kong, China: Association for Computational Linguistics, Nov. 2019, pp. 3585–3590. doi: 10.18653/v1/D19-1367.

**Appendix A. Search Strategy**

We used the following sources in the search:

1. PubMed (https://www.ncbi.nlm.nih.gov/pubmed/)

2. Web of Science (https://www.webofknowledge.com)

3. DBLP computer science bibliography (http://dblp.uni-trier.de/)

Based on the term clusters presented in Section 2.2 the search query was designed as follows:

(<de-identification> **OR** <de-identifying> **OR** <de-identify> **OR** <deidentif> **OR** <anonymi> **OR** <mask> **OR** <redact> **OR** <pseudonym>)

**AND**

(<clinical> **OR** <medical> **OR** <narrative> **OR** <discharge> **OR** <EHR> <**OR** <EMR> **OR** <EPR> **OR** <electronic health record> **OR** <electronic medical record> **OR** <electronic patient record> **OR** <electronic health records> **OR** <electronic medical records> **OR** <electronic patient records> **OR** <protected health information> **OR** <PHI> **OR** <text> **OR** <document >)

**AND**

(<natural language processing>**OR** <natural language processing> **OR** <NLP> **OR** <text mining> **OR** <i2b2> **OR** <named entity recognition> **OR** <NER> **OR** <information extraction>**OR** <dictionar> **OR** <machine learning>)

Following the specific syntax required by the databases we used following queries.

**A.1. Search in PubMed**

The search query used for PubMed is given below ([mh] denotes searching by MeSH term).

("Data Anonymization" [mh] **OR** "de-identification" [Title/Abstract] **OR** "de-identifying" [Title/Abstract] **OR** "de-identify" [Title/Abstract] **OR** deidentif* [Title/Abstract] **OR** anonymi*[Title/Abstract] **OR** mask* [Title/Abstract] **OR** redact* [Title/Abstract] **OR** pseudonym* [Title/Abstract] **OR** "i2b2/UTHealth" [Title/Abstract] **OR** "CEGS N-GRID" [Title/Abstract]) **AND** (clinical [Title/Abstract] **OR** medical [Title/Abstract] **OR** narrative* [Title/Abstract] **OR** discharge [Title/Abstract] **OR** EHR [Title/Abstract] **OR** EMR [Title/Abstract] **OR** EPR [Title/Abstract] **OR** "electronic health record" [Title/Abstract] **OR** "electronic medical record" [Title/Abstract] **OR** "electronic patient record" [Title/Abstract] OR "electronic health records" [mh] **OR** "electronic health records" [Title/Abstract] **OR** "electronic medical records" [Title/Abstract] **OR** "electronic patient records" [Title/Abstract] OR "protected health information" [Title/Abstract] **OR** PHI [Title/Abstract] OR text [Title/Abstract] OR document [Title/Abstract]) **AND** ("natural language processing" [mh] OR "natural language processing" [Title/Abstract] **OR** NLP **OR** "text mining" [Title/Abstract] OR "named entity recognition" [Title/Abstract] **OR** NER [Title/Abstract] **OR** "information extraction" [mh] **OR** "information extraction" [Title/Abstract] **OR** dictionar* [Title/Abstract] **OR** "machine learning" [mh] **OR** "machine learning" [Title/Abstract])

**A.2. Search in Web of Science**

The search query used for Web of science is given below.

("de-identification" **OR** "de-identifying" **OR** "de-identify" **OR** deidentif* **OR** anonymi* **OR** mask* **OR** redact* **OR** pseudonym* **OR** “i2b2/UTHealth” **OR** “CEGS N-GRID“) **AND** (clinical **OR** medical **OR** narrative* **OR** discharge **OR** EHR **OR** EMR **OR** EPR **OR** "electronic health record" **OR** "electronic medical record" **OR** "electronic patient record" **OR** "electronic health records" **OR** "electronic medical records" **OR** "electronic patient records" **OR** "protected health information" **OR** PHI **OR** text **OR** document ) **AND** ("natural language processing" **OR** NLP **OR** "text mining" **OR** i2b2 **OR** "named entity recognition" **OR** NER **OR** "information extraction" **OR** dictionar* **OR** “machine learning” )

**A.2. Search in DBLP**

The DBLP online documentation[5] states that, due to technical problems, the use of the dash (‘-‘) character is currently disabled, as well as that phrase search queries are not supported. Having that in mind, and the fact that abstracts cannot be searched through DBLP we adjusted our query as follows (the ‘|’ denotes the logical OR, while logical AND is accomplished using parenthesis):

(identif|deidentif|anonym|mask|redact|pseudonym|remov|detect)(clinical|medical|protected|PHI)(narrative|discharge|note|electronic|health|patient|record|EHR|EMR|EPR|text|document)

[5] https://dblp.uni-trier.de/faq/1474589.html

# Appendix B. Characteristics of the Included Studies

| Study | Task | Types of taggers | Features used by the ML model | | | | | | | | Methodology | | | | | Evaluation | | | | |
|---|---|---|---|---|---|---|---|---|---|---|---|---|---|---|---|---|---|---|---|---|
| | | | Lexical | Dictionary | Orthographic | Positional | Rules | Semantic | Word representation | Deep learning-based | Additional Information | Tools | Dictionary-based tagger | Machine learning model | Rule-based tagger | Precision | Recall | F-score | Matching strategy | Corpora |
| [26] | Systematic evaluation of feature impact | ML/Rules | ✓ | ✓ | ✓ | | ✓ | | | | rules only as features | CRFSuite | | CRF (BIO) | regex | 96.39 (2006 i2b2) | 98.49 (2006 i2b2) | 97.30 (2006 i2b2) | token level, micro | 2006 i2b2, PhysioNet |
| [24] | NER | ML | | | | | | | | | CRF features not reported, VMUC training data annotated automatically with DE-ID | Carafe, DE-ID | | CRF | | 94.30 (VUMC) 97.80 (2006 i2b2) | 97.80 (VUMC) 95.10 (2006 i2b2) | 96.00 (VUMC) 96.50 (2006 i2b2) | token level, micro | 2006 i2b2, VUMC |
| [69] | NER | ML/Dict/Rules | ✓ | ✓ | ✓ | | | | | | SVM with LVG features for false positive filtering | cTAKES, Lucene, Stanford NER re-trainable CRF, LIBSVM | fuzzy matching | CRF, SVM | | 77.40 | 97.40 | 92.60 (F2) | not reported | VHA |
| [36] | NER | ML/Dict/Rules | ✓ | | | | | | | | custom sentence splitter, false positive filtering - SVM | cTAKES, Lucene, Stanford NER CRF, LIBSVM | hard matching | CRF, SVM | regex | 83.60 (VHA) 84.60 | 92.20 (VHA) 96.50 | 87.70 (VHA) 90.20 | 'fully contained' level, micro | 2006 i2b2, VHA |
| [71] | NER | Dict/Rules | | | | | | | | | custom tokenizer, use of EHR structured data | Microsoft FAST | hard matching | | rules | 100.00 | 88.50 | not reported | not reported | SLaM |
| [70] | Experiments with active learning | ML | | | | | | | | | expansion of training data with manually corrected ML predictions | MITRE Identification Scrubber Toolkit (MIST) | | | | 97.30 | 95.50 | 96.40 | not reported | UMHS |
| [31] | NER | ML/Rules | ✓ | ✓ | | | ✓ | ✓ | | | | Weka | | J48 decision tree | rules based on term frequency | 61.00 | 98.00 | 75.00 | not reported | 2006 i2b2 |
| [25] | NER | ML/Dict/Rules | ✓ | ✓ | ✓ | | | | | | custom tokenizer | TreeTagger, MALLET | hard matching | CRF | regex | 95.73 (CCHMC) 99.18 (2006 i2b2) 89.02 (PhysioNet) | 92.91 (CCHMC) 94.26 (2006 i2b2) 58.49 (PhysioNet) | 94.30 (CCHMC) 96.68 (2006 i2b2) 70.60 (PhysioNet) | token level, micro | 2006 i2b2, PhysioNet, CCHMC |
| [89] | NER | Dict | | | | | | | | | created extensive dictionary of Patient and Provider names | | hard matching | | Deterministic Finite State Automaton (DFSA) | not reported | not reported | 75.60 | not reported | NIHCC |
| [72] | NER | Rules | | | | | | | | | | dTagger | | | regex, dTagger address patterns | 99.80 (specificity) | 99.40 (sensitivity) | 88.50 (F2) | not reported | NIHCC |
| [35] | NER | ML | ✓ | | ✓ | ✓ | ✓ | | | | | Stanford POS tagger | | CRF | | 98.99 (i2b2) | 88.59 (i2b2) | 93.00 (i2b2) 82.50 (CINSW) 98.07 (MTSamples) | not reported, macro | 2006 i2b2, MTSamples, CINSW |
| [29] | Experiments with de-id models trained on document clusters | ML | | | | | | | | | | MITRE Identification Scrubber Toolkit (MIST) | | | | not reported | not reported | 91.70 (VUMC) 96.60 (i2b2) | entity level, strict, macro | 2006 i2b2, VUMC |
| [43] | NER | ML | ✓ | ✓ | ✓ | | | | | | custom tokenizer | OpenNLP, Genia Tagger, CRF++ | | CRF (BIO) | | 95.61 | 89.25 | 92.32 | entity level, strict, micro | 2014 i2b2/UTHealth |
| [16] | NER | ML/Rules | ✓ | ✓ | ✓ | ✓ | | | ✓ | | Brown clustering, word2vec, merging strategy, MedEx tokenizer | Stanford NER, CRFsuite | | CRF (BIO and BIOES) | regex | 92.64 | 89.88 | 91.24 | entity level, strict, micro | 2014 i2b2/UTHealth |
| [51] | NER | ML/Rules | ✓ | ✓ | | | ✓ | | | | second pass tagging | Genia Tagger, CRF++ | | CRF (BIO) | regex | 96.45 | 90.92 | 93.60 | entity level, strict, micro | 2014 i2b2/UTHealth |
| [42] | NER | ML/Dict/Rules | ✓ | ✓ | ✓ | ✓ | | | | | second pass tagging, merging strategy | cTAKES, GATE, CRF++ | hard matching | CRF (IO) | regex, JAPE | 93.06 | 88.36 | 90.65 | entity level, strict, micro | 2014 i2b2/UTHealth |
| [41] | NER | ML/Rules | ✓ | ✓ | ✓ | | | | | | feature functions for out-of-vocabulary words | | | HMM, skip-chain CRF | rules - Python | 96.00 | 91.00 | 93.70 | token level, micro | 2014 i2b2/UTHealth |
| [73] | NER | Rules | | | | | | | | | | | | | regex | 81.30 | 66.50 | 73.10 | not reported | VA |
| [74] | NER | ML/Rules | | | | | ✓ | ✓ | | | cascaded system, use of EHR structured data, rules only as features | | | Logistic regression | regex | 96.00 | 94.00 | 95.00 | not reported, macro | NORC |
| [32] | Experiments with self-training | ML | ✓ | | | ✓ | | | | | | Stanford CoreNLP, CRFSharp | | CRF (BIO) | error correction rules | 97.63 | 93.15 | 95.33 | not reported | 2006 i2b2 |
| [33] | Experiments with self-training | ML/Rules | ✓ | | | ✓ | | | | | | Stanford CoreNLP, CRFSharp | | CRF (BIO) | error correction rules | 97.91 | 94.16 | 96.00 | not reported | 2006 i2b2 |
| [28] | NER | ML | | | | | | | ✓ | ✓ | text "skeleton" features | | | LSTM, GRU, (multi-BIO) | | 98.70 (2006) 99.31 (2014) | 98.62 (2006) 96.76 (2014) | 98.66 (2006) 98.02 (2014) | token level, micro | 2006 i2b2, 2014 i2b2/UTHealth |
| [49] | Evaluation of different word embeddings | ML | | | | | | | ✓ | ✓ | embeddings: random number initialization, RNN word embedding, CBOW | | | RNN (BIO) | | 97.26 | 90.67 | 93.84 | entity level, strict, micro | 2014 i2b2/UTHealth |
| [50] | NER | ML | | | | | | | ✓ | ✓ | CBOW embeddings | Stanford CoreNLP tokenizer | | RNN (BIO) | | 89.63 | 90.73 | 90.18 | entity level, strict, micro | 2014 i2b2/UTHealth |
| [67] | NER | ML | | ✓ | | | ✓ | | ✓ | ✓ | LSTM augmented with EHR structured and manually engineered features | WordNet | | bidirectional LSTM-CRF (BIO) | | 99.213 | 99.306 | 99.259 | binary token HIPPA, micro | MIMIC-III-D |
| [75] | NER | ML/Dict/Rules | ✓ | ✓ | ✓ | | | | | | created large-scale dictionary of Occupations | GATE, OpenNLP | hard matching | CRF (BIO) | rules | 88.34 (Christie) 88.79 (2014 i2b2/UTHealth) | 74.34 (Christie) 57.54 (ib2b 2014) | 80.74 (Christie) 69.83 (2014 i2b2/UTHealth) | entity level, strict, micro | 2014 i2b2/UTHealth, CNHS |
| [60] | NER | ML/Dict/Rules | | | | | | | | | | Stanford NER CRF | hard matching | CRF | regex and dictionary disambiguation rules | 91.62 | 83.38 | 87.31 | entity level, strict, micro | 2016 CEGS-NGRID |
| [46] | NER | ML/Rules | ✓ | ✓ | ✓ | | | ✓ | ✓ | ✓ | custom tokenizer, Brown clustering, word2vec, SVM based ensemble method | CRFsuite, LibSVM | | CRF, bidirectional LSTM-CRF (BIO) | regex | 96.46 (2014) 94.22 (2016) | 93.80 (2014) 88.81 (2016) | 95.11 (2014) 91.43 (2016) | entity level, strict, micro | 2014 i2b2/UTHealth, 2016 CEGS-NGRID |
| [61] | NER | ML/Dict/Rules | ✓ | ✓ | ✓ | | | | | | second pass tagging, merging strategy, addition of 2014 i2b2/UTHealth training data | cTAKES, GATE, CRF++ | hard matching | CRF (IO) | regex, JAPE | | | 87.69 | entity level, strict, micro | 2016 CEGS-NGRID |
| [88] | NER | ML/Rules | ✓ | ✓ | ✓ | ✓ | | ✓ | | | custom tokenizer, Brown clustering, word2vec, random indexing, addition of 2014 i2b2/UTHealth training data | CLAMP, OpenNLP | | CRF (BIO) | regex and error correction rules | 93.39 | 88.23 | 90.74 | entity level, strict, micro | 2016 CEGS-NGRID |
| [62] | NER | ML/Rules | ✓ | ✓ | ✓ | | | ✓ | | ✓ | custom tokenizer, character embeddings, wor2vec | OpenNLP, CRF++ | | CRF, bidirectional LSTM-CRF (BIOEU) | | | | 89.86 | entity level, strict, micro | 2016 CEGS-NGRID |
| [63] | Experiments with domain adaptation methods | ML | ✓ | ✓ | | ✓ | | ✓ | | | custom tokenizer, semantic roles, Brown clustering, word2vec, random indexing, | CLAMP OpenNLP, Stanford NER, SENNA, GloVe, CRFsuite | | CRF (BIO) | | 93.46 (2016) | 87.53 (2016) | 90.40 (2016) | entity level, strict, micro | 2006 i2b2, 2014 i2b2/UTHealth , 2016 CEGS-NGRID |
| [4] | NER | Dict/Rules | | | | | | | | | custom tokenizer, use of EHR structured data | GATE, MedEx-UIMA | hard matching | | regex | 97.80 | 89.70 | 93.60 | not reported | CPFT |
| [30] | NER | ML/Rules | ✓ | ✓ | | ✓ | | | | | | | | CRF, SVM | rules | 89.85 | 90.94 | 90.39 | not reported | 2006 i2b2 |
| [15] | NER | ML | | | | | | | ✓ | ✓ | word and character embeddings | Stanford CoreNLP tokenizer, GloVE | | bidirectional LSTM-CRF (BIO) | | 97.920 (2014) 98.820 (MIMIC) | 97.835 (2014) 99.398 (MIMIC) | 97.877 (2014) 99.108 (MIMIC) | binary token HIPPA, micro | 2014 i2b2/UTHealth, MIMIC-III-D |
| [47] | NER | ML | | | | | | | ✓ | ✓ | custom tokenizer, word2vec, CNN for character embeddings | | | bidirectional LSTM-CRF (BIO) | | not reported | not reported | 94.37 | entity level, strict, micro | 2014 i2b2/UTHealth |
| [76] | NER | Dict/Rules | | | | | | | | | use of EHR structured data | Stanford CoreNLP | hard matching | | regex | 99.50 (specificity) | 99.80 (sensitivity) | 86.90 | not reported | TAMHSC |
| [34] | NER | ML | | ✓ | | | | ✓ | ✓ | ✓ | | Keras | | bidirectional GRU (BIO) | | 99.03 (2006) 98.89 (2006) | 98.55 (2006) 97.23 (2014) | 98.79 (2006) 98.05 (2014) | token level, micro | 2006 i2b2, 2014 i2b2/UTHealth |
| [91] | Evaluation of binary vs. multi-class PHI classification | ML | ✓ | ✓ | ✓ | ✓ | | | | | | PTBTokenizer, Stanford NER re-trainable CRF | | CRF (IO) | | 89.90 (mulit-class) 98.50 | 97.00 (mulit-class) 92.00 | 93.30 (mulit-class) 95.10 | entity level, strict, micro | 2014 i2b2/UTHealth |
| [44] | NER | ML/Dict/Rules | | | | | | | ✓ | ✓ | ensemble of 12 taggers, custom tokenizer, word2vec | Stanford CoreNLP, NeuroNER, GloVE, Wapiti, Miralium, Vowpal Wabbi, MITIE, LIBLINEAR, MIST, PhysioNet deid | | CRF, MEMM, MIRA, SEARN, SVM, LSTM , LSTM-CRF (BIO) | | 97.04 | 94.45 | 95.73 | entity level, strict, micro | 2014 i2b2/UTHealth |
| [48] | NER | ML | ✓ | ✓ | | | | | ✓ | ✓ | custom tokenizer wor2vec | NLTK, GloVE sklearn-crfsuite, Keras | | CRF, bidirectional LSTM-CRF (BIOEU) | | not reported | not reported | 91.25 (CRF) 95.92 (LSTM) | token level, micro | 2014 i2b2/UTHealth |
| [78] | Experiments with active learning | ML | ✓ | | | | | | | | | MITRE Identification Scrubber Toolkit (MIST) | | | | not reported | not reported | 90.00 | entity level, strict, micro | 2014 i2b2/UTHealth |
| [92] | Experiments with fine-tuning of ML models on different corpora | ML | ✓ | ✓ | | | ✓ | | | ✓ | custom tokenizer, LSTM augmented with char. emendings and lex., synt, ort, and dict. features | TensorFlow | | LSTM-CRF (BIO) | | 94.74 | 91.09 | 92.88 | entity level, strict, micro | 2014 i2b2/UTHealth – test and fine-tuning, UFHealth - training |
| [45] | Evaluation of contextualized word embeddings | ML | | | | | | | ✓ | ✓ | | SpaCy, NeuroNER | | bidirectional LSTM- CRF, GRU | | 94.33 (2014) 90.88 (2016) | 91.36 (2014) 87.20 (2016) | 92.82 (2014) 89.00 (2016) | entity level, strict, micro | 2014 i2b2/UTHealth, 2016 CEGS-NGRID |
| [77] | NER | ML/Rules | | | | | | | | | de-id framework second pass, priority sorting | OpenNLP, Stanford CoreNLP, PRIMA, SystemT, Advance Care Insights (ACI) | | | | not reported | not reported | overall not reported, per category ~90.00 | not reported | USHP |
| [10] | Experiments with pseudonymised embeddings | ML | | | | | | | ✓ | ✓ | | SpaCy | | DANN | | not reported | not reported | 97.40 | binary token HIPPA, micro | 2014 i2b2/UTHealth |
| [27] | Experiments with fine-tuning of ML models on different corpora | ML | | ✓ | ✓ | | | | | | LSTM augmented with char. emendings and orthographic and affix features | vToken, GloVE | | bidirectional LSTM-CRF (BIO) | | 99.10 (2014) 99.60 (2006) 97.10 (mimic) | 85.70 (2014) 90.70 (2006) 96.30 (mimic) | 91.70 (2014) 94.90 (2006) 96.70 (mimic) | entity level, strict, micro | 2006 i2b2, 2014 i2b2/UTHealth, PhysioNet, MIMIC-III-H |
| [17] | NER | ML | | | | | | | | | deep contextualised embeddings – BERT, ELMo | | | bidirectional LSTM-CRF (BIO) | | 95.99 (2014) 93.39 (2016) | 95.02 (2014) 90.31 (2016) | 95.50 (2014) 91.82 (2016) | entity level, strict, micro | 2014 i2b2/UTHEalth, 2016 CEGS-NGRID |
| [19] | NER | ML | | | | | | | | ✓ | | | | BERT, SciBERT, BioBERT | | 98.98 (2014) 99.20 (2006) 95.61 (physio) 97.25 (mimic) | 98.27 (2014) 97.71 (2006) 95.61 (physio) 97.59 (mimic) | 98.62 (2014) 98.45 (2006) 95.61 (physio) 97.42 (mimic) | token level, binary | 2006 i2b2, 2014 i2b2/UTHealth, PhysioNet, MIMIC-III-D |
| [11] | Obfuscation using word embeddings | ML | | | | | | | ✓ | | replacement of every token with random closest N (3-14) tokens in the embedding space | FastTex for OOV words during evaluation | | CBOW, Skipgram used to train and replace embeddings | | | | authors argue against using P, R and F-score as evaluative measures for this task. They use semantic similarity (Pearson correlation) | every word | ICES, MIMIC-III-J for extrinsic evaluation |
| [53] | NER | ML/Rules | | | | | | | ✓ | ✓ | ensamble of 12 (22) de-id methods, GloVe embeddings used for all Bi-LSTM models, MetaMap used to recognize concepts and compare their occurrences across the two corpora | NeuroNER, PhysioNet deid, MITIE, MIST, LIBLINEAR, Wapiti, Vowpal Wabbit, Miralium | | LSTM, LSTM-CRF, CRF, MEMM, SEARN, MIRA, SVM, structured SVM, OGD, MIST (BIO) | PhysioNet deid | 95.45 (SVM-based stacked ensemble on 2016 N-GRID) | 90.08 (SVM-based stacked ensemble on 2016 N-GRID) | 92.69 (SVM-based stacked ensemble on 2016 N-GRID) | entity level, strict, binary token level | 2014 i2b2/UTHealth, 2016 CEGS-NGRID |
| [40] | NER | ML | | | | | | | ✓ | ✓ | bidirectional GRU for character-level embeddings, GloVe for word embeddings, CRF and softmax as classification layers, measured utility (BLEU, topic modeling, classification) and speed, 110,000,000 parameters | GloVe | | GRU, stacked RNNs (GRU-GRU, LSTM-GRU), BERT-like self-attention model (BIOES) | | 98.031 (self-attention, 2014) 99.957 (MIMIC-III) 89.20 (MIMIC-II) | 98.410 (2014) 98.788 (MIMIC-III) 82.90 (MIMIC-II) | 98.220 (2014) 99.369 (MIMIC-III) 85.90 (MIMIC-II) | binary token level | 2014 i2b2/UTHealth, MIMIC-III-T, MIMIC-II (PhysioNet) |
| [54] | NER | ML | | | | | | | | ✓ | DistilBERT (cased and uncased) was compared to ClinicalBERT and BlueBERT in terms of F1-score and speed | | | DistilBERT (BIO) | | | | 94.85 (DistilBERT, cased) 86.44 (DistilBERT, uncased) | entity level, strict | 2014 i2b2/UTHealth |
| [79] | Evaluation of off-the-shelf de-identification tools | ML/Dict/Rules | | | ✓ | | | | ✓ | ✓ | tokenization using SpaCy | SpaCy | MIT deid, NLM scrubber | MIST, Emory HIDE, NeuroNER | NLM-Scrubber, MIT deid | 94.5 (NeuroNER) | 92.6 (NeuroNER) | 93.6 (NeuroNER) | token level | UPHS-S |
| [37] | NER with domain adaptation experiments | ML | | | ✓ | | | | ✓ | ✓ | tokenization using spaCy, GloVe, BERT, fine-tuned GloVe embeddings, BiLSTM for character embeddings, hand-crafted embeddings for casing, used data augmentation (character and word-level switchout) and activation dropout for noise | GloVe, spaCy | | LSTM-CRF (Flair NER) | | | | 97.15 (2006, fully supervised) 94.01 (2014, fully supervised) 98.21 (mimic-discharge, fully supervised) 97.83 (mimic-ragiology, fully supervised) | strict entity level, micro | 2014 i2b2/UTHealth, 2006 i2b2, MIMIC-III-J |
| [55] | NER | ML | | | | | | | ✓ | ✓ | custom entity spacer, spaCy tokenizer, concatenated Flair and GloVe embeddings (character, word-level and contextual embeddings), sentence-grouping factor for wider context. Compared to BERT, MLR and MLR+CRF | NeuroNER, spaCy | | BiLSTM+CRF (IOB) | | 0.9653 (entity level) | 0.9506 (entity level) | 0.9579 (entity level) | entity, token, binary, micro | 2014 i2b2/UTHealth |
| [56] | NER | ML | | | | | | | | ✓ | sequence to sequence learning (enncoder-decoder architecture), 78,000,000 parameters | | | BERT | | 98.12 | 98.91 | 98.51 | binary token level | 2014 i2b2/UTHealth |
| [57] | NER | ML | | | | | | | ✓ | ✓ | custom tokenizer, character embeddings through char2vec, token embeddings | char2vec | | BiLSTM BiLSTM-CRF C2V-BiLSTM-CRF | | 83.91 (BiLSTM-CRF) | 81.80 (BiLSTM-CRF) | 82.84 (BiLSTM-CRF) | strict entity level | 2014 i2b2/UTHealth, RIH |

| | | | | | | | | | | | | | | | | | | | |
|---|---|---|---|---|---|---|---|---|---|---|---|---|---|---|---|---|---|---|---|
| | | | | | | | | | | | | | Transformer Transformer-CRF Transformer-BiLSTM (BIO) | | | | | | |
| [81] | NER (name removal) | ML | ✓ | | | | | | ✓ | ✓ | word2vec skip-gram model (FastText method), contextual embedding as cooccurrence weighted sum of the word2vec embeddings within a context window of 3 words, set of landmarks. Compared to Stanford NER and Naïve Bayes. | NLTK tokenizer, word2vec | | MLP (multilayer perceptron classifier) | | 99.73 (2 reports) | 1.0 (2 reports) | 88.90 ± 4.60 (6 reports) | binary token level | HCR |
| [58] | NER | ML/Dict/Rules | | | | | | | | ✓ | one model version was fine-tuned on Mayo, the other on Mayo+2014, HIPS method and surrogate post-processing | | dictionary of locations and organizations for replacement, sentence inclusion list | ensemble of fine-tuned BERT models (IOB2) | regex, pattern and template matching, support from structured EHR data | 97.9 (2014) 96.7 (Mayo) | 99.2 (2014) 99.4 (Mayo) | 98.5 (2014) 97.9 (Mayo) | entity, ,binary token level | 2014 i2b2/UTHealth, MC |
| [82] | Finding the best-performing features for ML-based NER model | ML | ✓ | | ✓ | ✓ | ✓ | ✓ | ✓ | | 10 features were extracted, 4 found to give best results (n-gram, prefix-suffix, word embedding, word shape), explored minimum training set size | MIST for annotation, CLAMP for NER model | | CRF | | 0.93 (name, all features combined) | 0.91 (name, all features combined) | 0.92 (name, all features combined) | not reported | UMH-R |
| [38] | NER | ML | ✓ | | ✓ | | | | ✓ | ✓ | attention applied to concatinated embeddings (character embeddings, token embeddings, deep prefix features, and n-gram context embeddings) | NeuroNER, GloVe | | BiLSTM-CRF | | 97.5 (2006) 94.2 (2014) 91.2 (2016) | 95.5 (2006) 91.5 (2014) 88.2 (2016) | 96.5 (2006) 92.8 (2014) 89.8 (2016) | micro, entity | 2014 i2b2/UTHealth, 2006 i2b2, 2016 CEGS-NGRID |
| [20] | NER | ML | | | | | | | ✓ | ✓ | ensambles by majority voting and stacking upon three groups of models, tok2vec, word2vec and BERT embeddings | custom web annotation tool, spaCy, Flair, Stanza, tok2vec, word2vec | | BiLSTM-CRF (Stanza and Flair), CNN (spaCy) (BIO) LR, SVM and XGB for ensamble stacking | | 99.55 (AHDS) 96.90 (2014, strict) 99.43 (2014 binary) | 98.77 (AHDS) 95.59 (2014 strict) 97.86 (2014 binary) | 99.16 (AHDS) 96.24 (2014 strict) 98.64 (2014 binary) | strict, entity, micro, binary | AHDS, 2014 i2b2/UTHealth for comparison |
| [83] | NER | ML/Dict/Rules | | | | | | | | | information form structured EHRs for PHI seeding, REDCap de-id review and adjudication tools | MIST, Stanford NLP, OpenNLP | hard matching (whitelisting and blacklisting) | CRF | regex | 57 (entity, physician notes) 94 (binary token, physician notes) 74.8 (entity, nursing flowsheets) | 98.1 (entity-level, physician notes) 99.7 (binary token, physician notes) 84.5 (entity, nursing flowsheets) | 92.6 (OOB Stanford NLP) | entity, binary token level | KUMC/MCW |
| [84] | NER | ML | | | | | | ✓ | ✓ | ✓ | word embeddings (GloVe), context-based word embeddings (Flair), character-level word embeddings (CNN), semantic embeddings as one-hot vectors | GloVe, Flair | | BiLSTM-CRF (BIO) | | 95.96 (2014) 96.23 (IC, mixed-domain embeddings) | 93.84 (2014) 94.51 (IC, mixed-domain embeddings) | 94.89 (2014) 95.36 (IC, mixed-domain embeddings) | strict entity level | IC, 2014 i2b2//UTHealth |
| [85] | Finding the best-performing features for ML-based NER model | ML | ✓ | | ✓ | ✓ | ✓ | | ✓ | | 10 features were extracted, 4 found to give best results (n-gram, prefix-suffix, word embedding, word shape), explored minimum training set size | MIST for annotation, CLAMP for NER model | | CRF | | 0.81 (name, all features combined) | 0.80 (name, all features combined) | 0.81 (name, all features combined)) | not reported | UMH-P |
| [86] | NER | ML/Dict/Rules | | | | | | | | | | Presidio, SpaCy | blacklisting | spaCy language model | regex | 89.21 | 80.64 (strict) 90.39 (relaxed) | 84.71 (strict) 89.80 (relaxed) | strict, relaxed | POWCC/POWCC MOSAIQ |
| [59] | Evaluation of different word embeddings | ML | | | | | | | ✓ | | word2vec, tokenization using spaCy, MWE dictionary applied during preprocessing, flashtext to join MWEs together | word2vec, spaCy, NLTK, flashtext | | CBOW, Skipgram used to replace embeddings | | | | 78% notes deidentified (UPHS-C, Wiki embeddings) 76% (UPHS-C, UPHS-C embeddings) 23.8% (2014, Wiki) 24.8% (2014) | patien's name and age (token level) | UPHS-C, 2014 i2b2//UTHealth for qualitative evaluation (8 notes) |
| [18] | NER | ML/Rules | | | | | | | | ✓ | HIPS algorithm for data augmentation, ULMFiT fine-tuning methods, Tree of Parzen estimator hyperparameter optimization | MIST, NLM-Scrubber, Emory HIDE, MIT deid, NeuroNER, transformer-based MIT deid (for comparison) | | PubMedBERT | hard-coded rules, stochastic approach to remove and replace PHI (HIPS algoritm) | 99.4 (2006) 98.6 (2014) | 99.5 (2006) 99.3 (2014) | 99.5 (2006) 98.9 (2014) | entity, binary token level | 2006 i2b2, 2014 i2b2//UTHealth , UPHS-R, UPHS-S, SMC |

**Appendix C.** Methods used for each PHI category**.**

| System | Age | Contact information | Date | Geographical location | Hospital | Identifiers | Person name | Profession |
|---|---|---|---|---|---|---|---|---|
| [26] | CRF | CRF | CRF | CRF+dictionary | CRF+dictionary | CRF | CRF+dictionary | – |
| [24] | CRF | CRF | CRF | CRF | CRF | CRF | CRF | – |
| [69] | – | – | – | – | – | – | CRF+rules+dictionary | – |
| [36] | Rules | Rules | CRF+rules | CRF+rules+dictionary | CRF+rules | Rules | CRF+rules | – |
| [71] | – | Rules+dictionary | Rules+dictionary | Rules+dictionary | Rules+dictionary | Rules+dictionary | Rules+dictionary | – |
| [70] | CRF | CRF | CRF | CRF | CRF | CRF | CRF | |
| [31] | J48 | J48 | J48 | J48 | J48 | J48 | J48 +dictionary | – |
| [25] | CRF | CRF+rules | CRF | CRF | CRF | CRF | CRF+dictionary | – |
| [89] | – | – | – | – | – | – | Rules+dictionary | – |
| [35] | CRF | CRF | CRF | CRF | CRF | CRF | CRF | – |
| [29] | CRF | CRF | CRF | CRF+dictionary | CRF+dictionary | CRF | CRF+dictionary | – |
| [72] | Rules | – | Rules | Rules | – | Rules | – | – |
| [43] | CRF | CRF | CRF +dictionary | CRF +dictionary | CRF | CRF | CRF | CRF |
| [16] | CRF | CRF+rules | CRF | CRF+dictionary | CRF | CRF+rules | CRF | CRF |
| [51] | CRF+rules | CRF+rules | CRF+rules+dictionary | CRF+rules+dictionary | CRF | CRF+rules | CRF+dictionary | CRF |
| [42] | Rules | Rules | CRF Rules | CRF+rules+dictionary | CRF +dictionary | Rules | CRF+rules | CRF +dictionary |
| [41] | HMM+CRF | HMM CRF+rules | HMM+CRF | HMM+CRF+rules | HMM+CRF | HMM+CRF+rules | HMM CRF+rules | HMM+CRF |
| [73] | Rules | Rules | Rules | Rules | Rules | Rules | Rules | – |
| [74] | – | Log. reg+rules+dictionary | – | Log. reg.+rules+dictionary | – | Log. reg.+rules+dictionary | Log. reg.+rules+dictionary | – |
| [32] | – | CRF +rules | CRF +rules | CRF +rules | CRF +rules | CRF | CRF +rules | – |
| [28] | LSTM | LSTM | LSTM | LSTM | LSTM | LSTM | LSTM | – |
| [49] | RNN | RNN | RNN | RNN | RNN | RNN | RNN | RNN |
| [50] | RNN | RNN | RNN | RNN | RNN | RNN | RNN | – |
| [67] | LSTM | LSTM | LSTM | LSTM | LSTM | LSTM | LSTM | – |
| [75] | – | – | – | – | – | – | CRF+rules+dictionary | – |
| [33] | CRF | CRF+rules | CRF+dictionary | CRF+rules+dictionary | CRF+rules | CRF+rules | CRF+rules | – |
| [60] | CRF Rules | CRF+rules | CRF Rules | CRF+rules+dictionary | CRF Rules +dictionary | Rules + | Rules+dictionary | CRF+dictionary |
| [46] | CRF+LSTM | CRF+LSTM+rules | CRF+LSTM | CRF+LSTM+dictionary | CRF+LSTM | CRF+LSTM Rules | CRF+LSTM | CRF+LSTM |
| [61] | CRF+rules | Rules | CRF Rules | CRF+rules+dictionary | CRF +dictionary | Rules | CRF+rules | CRF +dictionary |
| [88] | CRF+rules | CRF+rules | CRF+rules | CRF+rules+dictionary | CRF | Rules | CRF+dictionary | CRF |
| [62] | CRF+LSTM | CRF+LSTM | CRF+LSTM | CRF+LSTM | CRF+LSTM | CRF+LSTM | CRF+LSTM | CRF+LSTM |
| [63] | CRF | CRF | CRF | CRF | CRF | CRF | CRF | – |
| [4] | – | Rules | Rules | Rules | – | Rules | Rules | – |
| [30] | CRF+SVM+rules | – | CRF+SVM+rules | – | – | – | CRF+SVM+rules | – |
| [15] | LSTM | LSTM | LSTM | LSTM | LSTM | LSTM | LSTM | LSTM |
| [47] | LSTM | LSTM | LSTM | LSTM | LSTM | LSTM | LSTM | LSTM |
| [76] | – | Rules | CRF | CRF+rules | – | Rules | CRF+rules+dictionary | – |
| [34] | GRU | GRU | GRU | GRU | GRU | GRU | GRU | – |
| [91] | CRF | CRF | CRF | CRF | CRF | CRF | CRF | CRF |
| [44] | Ensemble of 12 models | Ensemble of 12 models | Ensemble of 12 models | Ensemble of 12 models+dictionary | Ensemble of 12 models | Ensemble of 12 models | Ensemble of 12 models+dictionary | Ensemble of 12 models |
| [48] | LSTM | LSTM | LSTM | LSTM | LSTM | LSTM | LSTM | LSTM |
| [78] | CRF | CRF | CRF | CRF | CRF | CRF | CRF | – |
| [92] | LSTM | LSTM | LSTM | LSTM | LSTM | LSTM | LSTM | – |
| [45] | LSTM | LSTM | LSTM | LSTM | LSTM | LSTM | LSTM | LSTM |
| [10] | LSTM | LSTM | LSTM | LSTM | LSTM | LSTM | LSTM | LSTM |
| [77] | Rules | CRF+rules | CRF Rules | CRF+rules | Rules | CRF+rules | CRF+rules | Rules |
| [27] | LSTM | LSTM | LSTM | LSTM | LSTM | LSTM | LSTM | LSTM |
| [17] | LSTM | LSTM | LSTM | LSTM | LSTM | LSTM | LSTM | LSTM |
| [19] | BERT, SciBERT, BioBERT | BERT, SciBERT, BioBERT | BERT, SciBERT, BioBERT | BERT, SciBERT, BioBERT | BERT, SciBERT, BioBERT | BERT, SciBERT, BioBERT | BERT, SciBERT, BioBERT | BERT, SciBERT, BioBERT |
| [11] | CBOW, Skipgram | CBOW, Skipgram | CBOW, Skipgram | CBOW, Skipgram | CBOW, Skipgram | CBOW, Skipgram | CBOW, Skipgram | CBOW, Skipgram |
| [53] | Ensemble of 12 models | Ensemble of 12 models | Ensemble of 12 models | Ensemble of 12 models | Ensemble of 12 models | Ensemble of 12 models | Ensemble of 12 models | Ensemble of 12 models |
| [40] | GRU, stacked RNNs, self-attention | GRU, stacked RNNs, self-attention | GRU, stacked RNNs, self-attention | GRU, stacked RNNs, self-attention | GRU, stacked RNNs, self-attention | GRU, stacked RNNs, self-attention | GRU, stacked RNNs, self-attention | GRU, stacked RNNs, self-attention |
| [54] | DistilBERT, ClinicalBERT, BlueBERT | DistilBERT, ClinicalBERT, BlueBERT | DistilBERT, ClinicalBERT, BlueBERT | DistilBERT, ClinicalBERT, BlueBERT | DistilBERT, ClinicalBERT, BlueBERT | DistilBERT, ClinicalBERT, BlueBERT | DistilBERT, ClinicalBERT, BlueBERT | DistilBERT, ClinicalBERT, BlueBERT |
| [79] | – | OOB NER tools | OOB NER tools | OOB NER tools | OOB NER tools | OOB NER tools | OOB NER tools | – |
| [37] | – | – | – | LSTM | LSTM | – | LSTM | – |
| [55] | LSTM | LSTM | LSTM | LSTM | LSTM | LSTM | LSTM | LSTM |
| [56] | BERT | BERT | BERT | BERT | BERT | BERT | BERT | BERT |
| [57] | LSTM, transformer | LSTM, transformer | LSTM, transformer | LSTM, transformer | LSTM, transformer | LSTM, transformer | LSTM, transformer | LSTM, transformer |
| [81] | – | – | – | – | – | – | MLP, Stanford NER, Naïve Bayes | – |
| [58] | Rules | Rules | Rules | BERT emsembles+dictionary for replacement | BERT emsembles+dictionary for replacement | Rules | BERT emsembles | – |
| [82] | – | – | CRF | CRF | CRF | CRF | CRF | – |
| [38] | LSTM | LSTM | LSTM | LSTM | LSTM | LSTM | LSTM | LSTM |
| [20] | – | Ensemble of NER machine learning models (CNN, LSTM) | Ensemble of NER machine learning models (CNN, LSTM) | Ensemble of NER machine learning models (CNN, LSTM) | – | Ensemble of NER machine learning models (CNN, LSTM) | Ensemble of NER machine learning models (CNN, LSTM) | – |
| [83] | Rules | Rules | Rules | Rules+OOB NER tools (CRF)+Dictionary | OOB NER tools (CRF) | Rules | OOB NER tools (CRF)+Dictionary | – |
| [84] | LSTM | LSTM | LSTM | LSTM | LSTM | LSTM | LSTM | – |
| [85] | – | CRF | CRF | CRF | CRF | CRF | CRF | – |
| [86] | – | Rules, context | Rules | Rules+Dictionary+spaCy | Rules+Dictionary+spaCy | Rules | spaCy+custom logic and context | Dictionary |
| [59] | CBOW, Skipgram | – | – | – | – | – | CBOW, Skipgram | – |
| [18] | – | PubMedBERT | PubMedBERT | PubMedBERT | PubMedBERT | PubMedBERT | PubMedBERT | – |

**Appendix D.** Performance on the three most frequently used publicly available corpora (2006 i2b2, 2014 i2b2/UTHealth, 2016 CEGS-NGRID).

| System | Methodology | Corpus | All categories | | | | | | HIPPA required categories | | |
|---|---|---|---|---|---|---|---|---|---|---|---|
| | | | Entity-level | | | Token-level | | | Binary token level | | |
| | | | P | R | F1 | P | R | F1 | P | R | F1 |
| [26] | CRF+dictionary | 2006 i2b2 | | | | 96.30 | 98.40 | **97.30** | | | |
| [24] | CRF | 2006 i2b2 | | | | 97.80 | 95.10 | 96.50 | | | |
| [36] | CRF+SVM | 2006 i2b2 | 96.50 | 87.70 | 90.20 | | | | | | |
| [25] | CRF+dictionary+rules | 2006 i2b2 | | | | **99.18** | 94.26 | 96.68 | | | |
| [31] | J48 | 2006 i2b2 | 61.00 | 98.00 | 75.00 | | | | | | |
| [29] | CRF | 2006 i2b2 | | | | | | 96.60 | | | |
| [35] | CRF | 2006 i2b2 | 98.99 | 88.59 | 93.00 | | | | | | |
| [33] | CRF+dictionary+rules | 2006 i2b2 | | | | 97.91 | 94.16 | 96.00 | | | |
| [32] | CRF+dictionary+rules | 2006 i2b2 | | | | 97.63 | 93.51 | 95.39 | | | |
| [28] | Bi-LSTM+WE | 2006 i2b2 | | | | | | | 98.70 | 98.62 | 98.66 |
| [30] | CRF+SVM+rules | 2006 i2b2 | 89.85 | 90.94 | 90.39 | | | | | | |
| [34] | Bi-GRU+WE | 2006 i2b2 | | | 94.52 | | | 95.40 | 99.03 | 98.55 | 98.79 |
| [27] | Bi-LSTM-CRF+CE+WE | 2006 i2b2 | **99.60** | 90.70 | 94.90 | | | | | | |
| [19] | $BERT_{base}$, uncased | 2006 i2b2 | | | | | | | 99.20 | 97.71 | 98.45 |
| [37] | LSTM-CRF+WE, CE | 2006 i2b2 | | | 97.15 | | | | | | |
| [38] | BiLSTM-CRF+CE+WE+DE (n-gram)+attention | 2006 i2b2 | 97.5 | 95.5 | 96.5 | | | | | | |
| [38] | BiLSTM-CRF+CE+WE+DE (ELMo) | 2006 i2b2 | 97.4 | 96.1 | 96.7 | | | | | | |
| [18] | PubMedBERT+ augmentation | 2006 i2b2 | 99.0 | **99.0** | **99.0** | | | | **99.4** | **99.5** | **99.5** |
| [43] | CRF+dictionary | 2014 i2b2/UTHealth | 95.61 | 89.25 | 92.32 | 97.33 | 93.04 | 95.14 | | | |
| [16] | CRF+dictionary+rules | 2014 i2b2/UTHealth | 92.64 | 89.88 | 91.24 | 95.64 | 93.66 | 94.64 | | | |
| [51] | CRF+dictionary+rules | 2014 i2b2/UTHealth | 96.45 | 90.92 | 93.60 | 98.06 | 94.14 | 96.11 | | | |
| [42] | CRF+dictionary+rules | 2014 i2b2/UTHealth | 93.06 | 88.36 | 90.65 | 97.22 | 92.50 | 94.80 | | | |
| [41] | HHM+CRF+rules | 2014 i2b2/UTHealth | 96.60 | 91.00 | 93.70 | | | | | | |
| [28] | Bi-LSTM+WE | 2014 i2b2/UTHealth | | | | | | | 99.31 | 96.76 | 98.02 |
| [49] | RNN+WE | 2014 i2b2/UTHealth | 97.26* | 90.67* | 93.84* | | | | | | |
| [50] | RNN+WE | 2014 i2b2/UTHealth | 90.73* | 89.63* | 90.18* | | | | | | |
| [46] | Bi-LSTM-CRF+CE+WE+CRF+dictionary+rules | 2014 i2b2/UTHealth | 96.46 | 93.80 | 95.11 | 97.94 | 96.04 | 96.98 | 98.88 | 97.66 | 98.27 |
| [15] | Bi-LSTM-CRF+CE+WE | 2014 i2b2/UTHealth | | | | | | | 98.32 | 97.38 | 97.84 |
| [15] | Bi-LSTM-CRF+CE+WE+CRF | 2014 i2b2/UTHealth | | | | | | | 97.92 | 97.83 | 97.87 |
| [47] | Bi-LSTM-CRF+CE+WE | 2014 i2b2/UTHealth | | | 94.37 | | | | | | 98.05 |
| [34] | Bi-GRU+WE | 2014 i2b2/UTHealth | | | 93.44 | | | 94.01 | 98.89 | 97.23 | 98.05 |
| [44] | Ensemble of 12 taggers | 2014 i2b2/UTHealth | 97.04 | 94.45 | 95.73 | | | | 99.16 | 98.06 | 98.61 |
| [48] | Bi-LSTM-CRF+CE+WE | 2014 i2b2/UTHealth | | | | | | 95.92 | | | |
| [52] | Bi-LSTM-CRF+CE+WE | 2014 i2b2/UTHealth | 96.97 | 94.01 | 95.47 | | | | | | |
| [45] | Bi-LSTM-CRF+CE+WE+DE | 2014 i2b2/UTHealth | 94.33 | 91.36 | 92.82 | | | | | | |

| | | | | | | | | | | | | |
|---|---|---|---|---|---|---|---|---|---|---|---|---|
| [10] | Bi-LSTM-CRF+WE | 2014 i2b2/UTHealth | | | | | | | | | 97.40 |
| [27] | Bi-LSTM-CRF+CE+WE | 2014 i2b2/UTHealth | **99.10** | 85.70 | 91.70 | | | | | | 97.70 |
| [17] | Bi-LSTM-CRF+DE(BERT) | 2014 i2b2/UTHealth | 95.99 | 95.02 | 95.50 | 97.79 | 97.16 | 97.48 | 98.98 | 98.45 | 98.71 |
| [19] | BERT$_{base}$, uncased | 2014 i2b2/UTHealth | | | | 98.61 | 97.90 | 98.25 | 98.98 | 98.27 | 98.62 |
| [19] | BERT$_{large}$, uncased | 2014 i2b2/UTHealth | | | | 98.66 | **98.15** | **98.40** | 99.08 | 98.57 | 98.82 |
| [53] | Stacked SVM | 2014 i2b2/UTHealth | | | 95.70 | | | | | | |
| [40] | BERT-like self attention | 2014 i2b2/UTHealth | | | | | | | 98.03 | 98.41 | 98.22 |
| [40] | BERT-like self attention+CRF decoding layer | 2014 i2b2/UTHealth | | | | | | | 98.71 | 97.923 | 98.313 |
| [40] | GRU-GRU+CE+WE | 2014 i2b2/UTHealth | | | | | | | 99.01 | 95.124 | 97.028 |
| [54] | DistilBERT, cased | 2014 i2b2/UTHealth | | | 94.85 | | | | | | |
| [54] | ClinicalBERT, cased | 2014 i2b2/UTHealth | | | 95.38 | | | | | | |
| [37] | LSTM-CRF+WE, CE | 2014 i2b2/UTHealth | | | 94.01 | | | | | | |
| [55] | BiLSTM+CRF+CE+WE+DE (Flair) | 2014 i2b2/UTHealth | 96.53 | 95.06 | 95.79 | **98.70** | 97.16 | 97.92 | 99.10 | 97.55 | 98.32 |
| [56] | BERT | 2014 i2b2/UTHealth | | | | | | | 98.12 | 98.91 | 98.51 |
| [57] | BiLSTM+CRF+WE | 2014 i2b2/UTHealth | 83.91 | 81.80 | 82.84 | | | | | | |
| [58] | BERT ensembles+dictionary+rules | 2014 i2b2/UTHealth | | | | | | | 97.9 | 99.2 | 98.5 |
| [38] | BiLSTM-CRF+CE+WE+DE (n-gram)+attention | 2014 i2b2/UTHealth | 94.2 | 91.5 | 92.8 | | | | | | |
| [38] | BiLSTM-CRF+CE+WE+DE (ELMo) | 2014 i2b2/UTHealth | 94.7 | 91.8 | 93.3 | | | | | | |
| [20] | Ensemble of best three F1 NER models (LSTM, CNN) +WE+DE (BERT) | 2014 i2b2/UTHealth | 96.90 | **95.59** | **96.24** | | | | **99.43** | 97.86 | 98.64 |
| [84] | Bi-LSTM-CRF+CE+WE+DE (Flair) | 2014 i2b2/UTHealth | 95.96 | 93.84 | 94.89 | | | | | | |
| [18] | PubMedBERT+augmentation | 2014 i2b2/UTHealth | 96.1 | 92.5 | 93.9 | | | | 98.6 | **99.3** | **98.9** |
| [60] | CRF+dictionary+rules | 2016 CEGS-NGRID | 91.62 | 83.38 | 87.31 | 97.21 | 88.11 | 92.44 | 96.29 | 91.18 | 93.66 |
| [46] | Bi-LSTM-CRF+CE+WE+CRF+dictionary+rules | 2016 CEGS-NGRID | 94.22 | 88.81 | 91.43 | 97.78 | **92.81** | 93.07 | 97.06 | 93.05 | 95.01 |
| [61] | CRF+dictionary+rules | 2016 CEGS-NGRID | | | 85.72 | | | 88.57 | | | |
| [88] | CRF+dictionary+rules | 2016 CEGS-NGRID | 93.39 | 88.23 | 90.74 | **97.84** | 92.62 | **95.16** | **97.18** | 93.09 | 95.09 |
| [62] | Bi-LSTM-CRF+CE+WE+CRF | 2016 CEGS-NGRID | | | 89.86 | | | 92.18 | | | 93.51 |
| [45] | Bi-LSTM-CRF+CE+WE+DE | 2016 CEGS-NGRID | 90.88 | 87.20 | 89.00 | | | | | | |
| [17] | Bi-LSTM-CRF+DE(BERT) | 2016 CEGS-NGRID | 93.39 | **90.31** | 91.82 | 95.36 | 92.32 | 93.81 | 96.96 | **94.39** | **95.66** |
| [53] | Stacked SVM | 2016 CEGS-NGRID | **95.45** | 90.08 | **92.69** | | | | 98.57 | 92.86 | 95.63 |
| [38] | BiLSTM-CRF+CE+WE+DE (n-gram)+attention | 2016 CEGS-NGRID | 91.2 | 88.2 | 89.8 | | | | | | |
| [38] | BiLSTM-CRF+CE+WE+DE (ELMo) | 2016 CEGS-NGRID | 92.2 | 89.0 | 90.5 | | | | | | |

**P**-precision, **R**-recall, **F1**-F1score; **WE**, **CE**, **DE** - word, character and contextual embeddings; ***not evaluated on the same test set as other approaches.

## Appendix E.

Entity-level per category performance for the main categories defined in Table 2. Best performing approaches that reported per category results (for the 2014 i2b2/UTHealth and 2016 CEGS-NGRID corpora) grouped by three most typical system architectures.

| **2014 i2b2/UTHealth** | **Yang and Garibaldi** [51] **CRF** | | | **Liu et al.** [16] **CRF+RNN** | | | **Tang et al.** [17] **RNN** | | | **Chambon et al.** [18] **PubMedBERT** | | | **Number of instances** | | |
|---|---|---|---|---|---|---|---|---|---|---|---|---|---|---|---|
| **Category** | **Precision** | **Recall** | **F1-score** | **Precision** | **Recall** | **F1-score** | **Precision** | **Recall** | **F1-score** | **Precision** | **Recall** | **F1-score** | **Train** | **Test** | **Total** |
| ***AGE*** | 97.12 | 92.54 | 94.77 | 98.66 | 96.34 | 97.48 | 97.51 | 97.38 | 97.45 | - | - | - | 1233 | 764 | 1997 |
| ***CONTACT*** | 96.63 | 92.20 | 94.37 | 97.65 | 95.41 | 96.52 | 96.36 | 97.25 | 96.8 | 99.0 | 98.8 | 98.9 | 323 | 218 | 541 |
| ***DATE*** | 98.67 | 96.63 | 97.64 | 98.34 | 97.69 | 98.02 | 98.97 | 98.53 | 98.75 | 99.6 | 99.6 | 99.6 | 7495 | 4980 | 12475 |
| ***ID*** | 93.78 | 91.68 | 92.72 | 94.41 | 91.84 | 93.11 | 92.97 | 91.04 | 92.00 | 97.4 | 98.8 | 98.1 | 4456 | 2883 | 7339 |
| ***LOCATION*** | 90.61 | 76.12 | 82.73 | 92.66 | 85.00 | 88.67 | 90.47 | 86.93 | 88.66 | 92.7 | 95.8 | 94.3 | 881 | 625 | 1506 |
| ***NAME*** | 96.96 | 91.68 | 94.24 | 95.42 | 94.03 | 94.72 | 95.26 | 94.83 | 95.05 | 97.3 | 97.4 | 97.4 | 2767 | 1813 | 4580 |
| ***PROFESSION*** | 81.06 | 59.78 | 68.81 | 91.34 | 64.80 | 75.82 | 82.32 | 83.24 | 82.78 | 97.3 | 97.4 | 97.4 | 234 | 179 | 413 |
| **Average:** | 93.55 | 85.80 | 89.33 | 95.49 | 89.30 | 92.04 | 93.40 | 92.74 | 93.07 | 97.22 | 97.97 | 97.62 | | | |
| | | **Macro:** | 89.51 | | **Macro:** | 92.29 | | **Macro:** | 93.07 | | **Macro:** | 93.9 | | | |
| | | **Micro:** | 93.60 | | **Micro:** | 95.11 | | **Micro:** | 95.50 | | **Micro:** | - | | | |

| **2016 CEGS-NGRID** | | **Lee et al.** [88]<br>**CRF** | | | **Liu et al.** [46]<br>**CRF+RNN** | | | **Tang et al.** [17]<br>**RNN** | | | **Kim et el.** [53]<br>**stacked SVMs** | | | **Number of instances** | |
|---|---|---|---|---|---|---|---|---|---|---|---|---|---|---|---|
| **Category** | **Precision** | **Recall** | **F1-score** | **Precision** | **Recall** | **F1-score** | **Precision** | **Recall** | **F1-score** | **Precision** | **Recall** | **F1-score** | **Train** | **Test** | **Total** |
| ***AGE*** | 96.00 | 93.88 | 94.93 | 97.36 | 95.67 | 96.51 | 97.26 | 96.52 | 96.89 | 97.75 | 95.75 | 96.74 | 3637 | 2354 | 5991 |
| ***CONTACT*** | 93.50 | 91.27 | 92.37 | 92.74 | 91.27 | 92.00 | 91.94 | 90.48 | 91.20 | 94.78 | 86.51 | 90.46 | 154 | 126 | 280 |
| ***DATE*** | 97.04 | 95.32 | 96.17 | 96.99 | 96.05 | 96.52 | 97.34 | 96.73 | 97.03 | 97.54 | 96.57 | 97.05 | 5723 | 3821 | 9544 |
| ***ID*** | 95.24 | 60.61 | 74.07 | 81.82 | 54.55 | 65.45 | 58.82 | 60.61 | 59.70 | 79.17 | 57.58 | 66.67 | 44 | 33 | 77 |
| ***LOCATION*** | 88.47 | 81.36 | 84.76 | 89.87 | 80.48 | 84.92 | 88.06 | 82.76 | 85.33 | 92.54 | 82.92 | 87.47 | 213 | 3771 | 10984 |
| ***NAME*** | 94.48 | 92.47 | 93.46 | 96.20 | 91.60 | 93.84 | 94.99 | 92.30 | 93.63 | 95.98 | 93.43 | 94.69 | 3691 | 2404 | 6095 |
| ***PROFESSION*** | 86.44 | 64.36 | 73.78 | 85.61 | 70.69 | 77.44 | 85.03 | 75.94 | 80.23 | 90.27 | 72.57 | 80.46 | 1471 | 1010 | 2481 |
| **Average:** | 93.02 | 82.75 | 87.08 | 91.51 | 82.90 | 86.67 | 87.63 | 85.05 | 86.29 | 92.58 | 83.62 | 87.65 | | | |
| | | **Macro:** | 87.58 | | **Macro:** | 86.99 | | **Macro:** | 86.32 | | **Macro:** | 87.87 | | | |
| | | **Micro:** | 90.74 | | **Micro:** | 91.43 | | **Micro:** | 91.82 | | **Micro:** | 92.69 | | | |